%% file: main.tex
\documentclass{article}

\usepackage[preprint]{neurips_2026}

\usepackage[utf8]{inputenc} 
\usepackage[T1]{fontenc}    
\usepackage{hyperref}       
\usepackage{url}            
\usepackage{booktabs}       
\usepackage{amsfonts}       
\usepackage{nicefrac}       
\usepackage{microtype}      
\usepackage{xcolor}         
\usepackage{enumitem}
\usepackage{multirow}
\usepackage{makecell}
\usepackage{graphicx}   
\usepackage{subcaption} 
\usepackage{wrapfig}
\usepackage{caption}
\usepackage{colortbl}
\usepackage{amsmath, amsthm, amssymb}
\newtheorem{theorem}{Theorem}
\newtheorem{assumption}{Assumption}

\definecolor{gaingreen}{RGB}{34,139,34}
\definecolor{extrap}{RGB}{219,234,254}

\title{Linear Dynamics in the RLVR Training of Large Language Models}

\author{\small
\hspace{-0.3em}Tianle Wang\textsuperscript{1,2,3*}\hspace{0.3em}
Jiayu Liu\textsuperscript{2*}\hspace{0.3em}
Zhongyuan Wu\textsuperscript{3,4*}\hspace{0.3em}
Shenghao Jin\textsuperscript{3,4}\hspace{0.3em}
Wei Chen\textsuperscript{3}\hspace{0.3em} 
Hao Xu\textsuperscript{3}\hspace{0.3em}
Ning Miao\textsuperscript{1,2}
}

\begin{document}

\maketitle
\footnotetext[1]{Department of Data Science, City University of Hong Kong,~
\textsuperscript{2}Hong Kong Institute of AI for Science, City University of Hong Kong,~
\textsuperscript{3}Li Auto Inc.,~
\textsuperscript{4}Beihang University,~
\textsuperscript{*}Equal contribution. Correspondence to: Ning Miao (ningmiao@cityu.edu.hk), Hao Xu (kingsleyhsu1@gmail.com).}

\begin{abstract}
\input{sections/0_abstract}
\end{abstract}

\input{sections/1_intro}

\input{sections/2_related_work}

\input{sections/3_empirical_law}

\input{sections/4_origin}

\input{sections/5_extrapolation}

\input{sections/6_conclusion}

\bibliographystyle{plainnat} 

\bibliography{refs}
\medskip

\newpage
\appendix
\input{sections/7_appendix}



\end{document}

%% file: sections/0_abstract.tex
Reinforcement learning with verifiable rewards (RLVR) has driven significant performance gains in reasoning-oriented large language models (LLMs), yet its internal training dynamics remain largely a black box.
In this work, we perform a comprehensive trajectory-level analysis of RLVR and uncover a striking regularity: across various model families, RL algorithms, and training configurations, RLVR consistently enters a \textit{robust linear regime}, where both parameter weights and output log-probabilities, measured rigorously via teacher-forced evaluation, evolve in a highly linear manner ($R^2 > 0.7$).
Through controlled experiments and theoretical analysis, we demonstrate that this linearity is not a coincidence, but stems from the high-variance, noisy nature of RLVR training signals, which act as a low-pass filter to concentrate optimization along a stable, low-dimensional drift.
Moreover, we show that this linear structure is not merely descriptive but powerfully predictive and actionable.
Specifically, weight-space extrapolation matches the performance of standard RL optimization while achieving a 6.1× training speedup through periodic re-grounding.
Meanwhile, output-space extrapolation serves as a lightweight intervention that effectively bypasses late-stage model collapse, consistently outperforming standard RL across mathematical and coding benchmarks, with an average performance improvement of 4.2\%.
Our code is available at \url{https://github.com/Miaow-Lab/RLVR-Linearity}.

%% file: sections/1_intro.tex
\section{Introduction}
Reinforcement learning with verifiable rewards (RLVR) has emerged as a pivotal stage in the post-training of reasoning-oriented large language models (LLMs), leading to high performance gain in math and coding~\citep{jaech2024openai, lambert2024tulu, Guo_2025}.
However, the black-box nature of LLMs leaves their evolution during this stage largely unknown. A deeper understanding of RLVR training dynamics is thus essential, both to explain where the performance gains come from and to improve the controllability, stability, and efficiency of post-training.

Recent work has examined RLVR dynamics from several complementary perspectives, including exploration-exploitation~\citep{cheng2026reasoning, agarwal2025unreasonableeffectivenessentropyminimization}, parameter-space dynamics~\citep{NEURIPS2025_bf235a1d, zhu2025the}, and behavioral patterns~\citep{pmlr-v267-lin25j, gandhi2025cognitivebehaviorsenableselfimproving}.
Despite this growing body of work, they have been largely focused on either too microscopic per-step model updates or too macroscopic behavior changes after the whole training process, leaving the training path itself underexplored.
Per-step analyses can characterize individual weight updates, but they lack a global view of the whole training process and the correlation between training steps. For example, a specific pattern reinforced in an RLVR step might be suppressed in subsequent ones, causing their effects to ultimately cancel out or accumulate into a coherent stronger direction in long run.
On the other hand, macroscopic analysis only reveals the high-level changes in model behaviors---such as shifts in response length, or the emergence of self-correction behaviors---without understanding the internal mechanisms. 
Thus, several central questions in RLVR are left unanswered because of their inherent dependence on trajectory-level analysis, including where performance gains come from, or whether RLVR instills genuinely new reasoning capabilities~\citep{wen2025reinforcementlearningverifiablerewards, wang2026newskillssharperprimitives} or primarily elicits latent ones~\citep{wu2026invisibleleashrlvrescape, NEURIPS2025_537d5aa7}.
\begin{figure}[t]
    \centering
    \includegraphics[width=\textwidth]{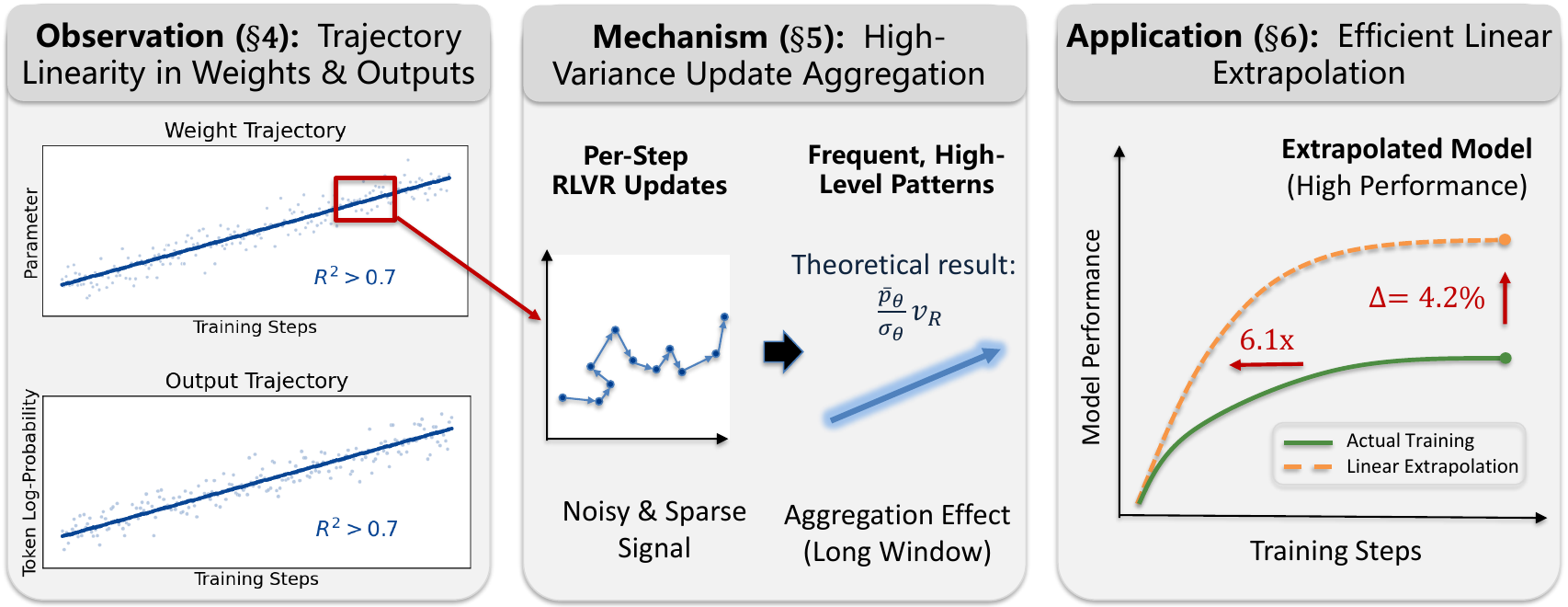}
    \caption{An overview of our main findings. In \S~\ref{sec:regime}, we find that model weights and outputs evolve linearly throughout RLVR training, consistently achieving a $R^2 > 0.7$. In \S~\ref{sec:origin}, we identify that the aggregation of noisy, high-variance RL updates is the core reason for this linearity. In \S~\ref{sec:extrapolation}, we show that linear extrapolation from early steps predicts future models, reducing training cost while preserving or improving performance.}
    \label{fig:example}
    \vspace{-10pt}
\end{figure}

In this work, we perform a comprehensive trajectory-level analysis of RLVR dynamics. Our empirical investigation reveals a striking phenomenon: the evolution of both LLM weights and outputs consistently exhibits a high degree of trajectory linearity. We characterize this as a \textit{robust linear regime} (§\ref{sec:regime}), which persists throughout the majority of the training process. Concretely, both weight and output trajectories achieve a coefficient of determination ($R^2$) exceeding 0.7.
This pattern is consistent across a broad range of base models (DeepSeek-R1-Distill-series~\citep{Guo_2025}, Open-Nemotron-1.5B~\citep{moshkov2025aimo2winningsolutionbuilding}, Qwen3-series~\citep{yang2025qwen3technicalreport}), RL algorithms (GRPO~\citep{shao2024deepseekmathpushinglimitsmathematical}, REINFORCE++~\citep{hu2025reinforcestabilizingcriticfreepolicy} and GSPO~\citep{zheng2025groupsequencepolicyoptimization}), and training setups, suggesting that RLVR trajectories are far more structured than their apparent complexity suggests.
The observed linearity implies a surprising degree of directional stability: rather than continually discovering new directions, RLVR enters a regime where later updates increasingly reinforce and amplify directions established earlier. This raises a natural puzzle:
\begin{center}
    \emph{Why does a highly nonlinear Transformer trained with stochastic RL exhibit stable linear dynamics in parameter and output space that are mechanistically grounded and practically predictive?}
\end{center}

We first provide a mechanistic account of how this linearity emerges (\S~\ref{sec:origin}).
We rule out several plausible explanations: e.g., that the small learning rate confines updates to a locally linear neighborhood, or that the AdamW optimizer smooths nonlinear fluctuations. Instead, we identify the core cause as the \emph{high-variance training signal} of RLVR. Specifically, because the signal is sparse and noisy, low-frequency patterns are extremely hard to learn. The model is therefore driven to update primarily along frequent patterns, which lie in a highly linear region of the loss landscape. This explains why per-step updates appear noisy yet aggregate into a stable linear drift over longer windows. Controlled experiments on supervised fine-tuning (SFT) further corroborate this account, indicating that the observed linearity is not a coincidence but a consequence of structured dynamics intrinsic to RLVR.

We further show that this linear structure is not only descriptive but also predictive (\S~\ref{sec:extrapolation}).  Specifically, a future model can be obtained by linearly extrapolating from existing training steps.
Experiments on multiple benchmarks (AIME24/25~\citep{aime24, aime25}, MATH500~\citep{math500}, and LiveCodeBench~\citep{livecodebench}) show that the extrapolated models attain performance very close to that of continued training, even when the extrapolation horizon is large.
For output-space extrapolation, the predicted model can even surpass the point of model collapse and outperform actually trained checkpoints.
The linear regime is thus mechanistically grounded enough to support accurate forecasting, which incidentally opens up new avenues for efficient post-training.

In summary, our contributions are as follows (see Figure~\ref{fig:example}):
\begin{itemize}[topsep=0pt,
  partopsep=0pt,
  itemsep=2pt,
  parsep=0pt,
  leftmargin=*]
    \item We uncover a trajectory-level regularity in RLVR: across model families, RL algorithms, and training setups, both parameter and token log-probability evolve linearly over training steps, with the majority of weights and tokens exhibiting $R^2 > 0.7$.
    \item We provide a mechanistic account for the linearity empirically and theoretically, demonstrating how the long-window aggregation of high-variance RL updates forms a stable drift direction.
    \item We show that this regime can be exploited, not just observed: weight-space extrapolation achieves standard RL performance with a \textbf{6.1x} speedup, while output-space extrapolation improves final performance by \textbf{4.2\%}.
\end{itemize}

%% file: sections/2_related_work.tex
\section{Related Work}
Recent studies have increasingly investigated the internal mechanisms of RLVR along two main threads: uncovering the structured parameter dynamics that drive reasoning gains, and exploiting the resulting linear weight space for model merging and extrapolation.

\paragraph{Parameter Dynamics in LLM Post-Training.}
Despite relying on full-parameter optimizers such as AdamW, recent work shows that RLVR weight updates are in fact tightly constrained along three axes: spatial, geometric, and temporal.
Spatially, \citet{NEURIPS2025_bf235a1d} observe extreme sparsity in RL fine-tuning, with updates largely confined to small subnetworks (roughly $20\%$ of parameters activated).
Geometrically, \citet{zhu2025the} formally prove that RLVR traverses the ``path not taken'', optimizing primarily along non-principal directions of the Hessian. This orthogonal perturbation explains why RLVR achieves strong reasoning gains with high sample efficiency~\citep{wang2026reinforcement} while averting catastrophic forgetting of pre-trained linguistic capabilities.
However, both spatial and geometric analyses remain strictly descriptive, offering structural insights without translating them into actionable training procedures.
Temporally, this structural rigidity manifests as ``Rank-1 Dominance''~\citep{cai2026predictabilityreinforcementlearningdynamics}, where the parameter update matrix ($\Delta W$) is overwhelmingly governed by its top singular subspace that dictates over 99\% of reasoning improvements and exhibits strictly linear dynamics throughout training. Yet, this study merely identifies rank-1 dominance as an empirical phenomenon without providing a corresponding mechanistic explanation.

Moreover, existing studies stop at weight-space artifacts and do not characterize their implications in output space. We address these gaps by mapping low-rank weight dynamics to output log-probabilities, providing a mechanistic account of how low-rank parameter dynamics deterministically propagate to token-level linear evolution and constrain capability acquisition under RLVR.

\paragraph{Model Merging and Extrapolation.}
A separate line of work exploits the linear weight space for checkpoint combination and extrapolation.
Methods such as Model Soup~\citep{wortsman2022modelsoupsaveragingweights} leverage the linear connectivity of fine-tuned models within low-error basins to perform spatial interpolation, while ExPO~\citep{zheng2025modelextrapolationexpeditesalignment} extrapolates between SFT and aligned checkpoints to address cross-stage distribution mismatch.
These methods, however, treat linear operability as a useful empirical property rather than asking why it holds. We focus specifically on the RLVR stage and provide a mechanistic explanation for why such simple linear operations on weight space succeed, which in turn enables principled performance prediction along the training trajectory.

\section{Preliminaries in RLVR}
Reinforcement Learning with Verifiable Rewards (RLVR) bypasses the noisy reward models of standard RLHF by optimizing language models against deterministic, rule-based feedback (e.g., mathematical correctness)~\citep{shao2024deepseekmathpushinglimitsmathematical, NEURIPS2022_8636419d}. To mitigate the memory overhead of the actor-critic architecture in standard PPO~\citep{schulman2017proximal}, modern RLVR heavily relies on critic-free algorithms.

A representative algorithm in this paradigm is Group Relative Policy Optimization (GRPO)~\citep{shao2024deepseekmathpushinglimitsmathematical}. For a policy $\pi_\theta$ and a given prompt $x$, GRPO samples a group of $G$ responses $\{y_i\}_{i=1}^G$ from the reference policy $\pi_{\theta_{\text{old}}}$. The importance weight for the $t$-th token is defined as $w_{i,t}(\theta) = \frac{\pi_{\theta}(y_{i,t} \mid x, y_{i,<t})}{ \pi_{\theta_{\text{old}}}(y_{i,t} \mid x, y_{i,<t})}$. 
Since $r(x, y_i)$ is a response-level reward, the advantage is uniform across tokens within each response and is computed by standardizing within the group:
\[
\hat{A}_{i,t} = \hat{A}_i = \frac{r(x, y_i) - \operatorname{mean}\bigl(\{r(x, y_i)\}_{i=1}^G\bigr)}{\operatorname{std}\bigl(\{r(x, y_i)\}_{i=1}^G\bigr)}.
\]
The GRPO objective integrates these group-normalized advantages with a PPO-style surrogate loss:
\[
\mathcal{J}_{\text{GRPO}}(\theta) = \mathbb{E}_{x \sim \mathcal{D}, \{y_i\}_{i=1}^G} \left[ \frac{1}{G} \sum_{i=1}^G \frac{1}{|y_i|} \sum_{t=1}^{|y_i|} \operatorname{min} \left( w_{i,t}(\theta) \hat{A}_{i,t},\; \operatorname{clip}\bigl(w_{i,t}(\theta), 1 \pm \epsilon\bigr) \hat{A}_{i,t} \right) \right].
\]
For simplicity in our trajectory analysis, we focus on the core surrogate objective, omitting the KL divergence penalty typically used to prevent policy drift.

%% file: sections/3_empirical_law.tex
\section{Empirical Phenomenon: RLVR Trajectories are Almost Linear}
\label{sec:regime}
In this section, we quantitatively demonstrate that RLVR trajectories exhibit striking linearity, both in the weight (parameter) space and the output-space.
To rule out the possibility that this phenomenon is an artifact of specific experimental configuration, we evaluate a total of 13 diverse settings, showing that this linearity persists robustly across various models, data regimes, and RL algorithms.

\subsection{Linearity in Weights}
\begin{figure}[t]
    \centering
    \includegraphics[width=\textwidth]{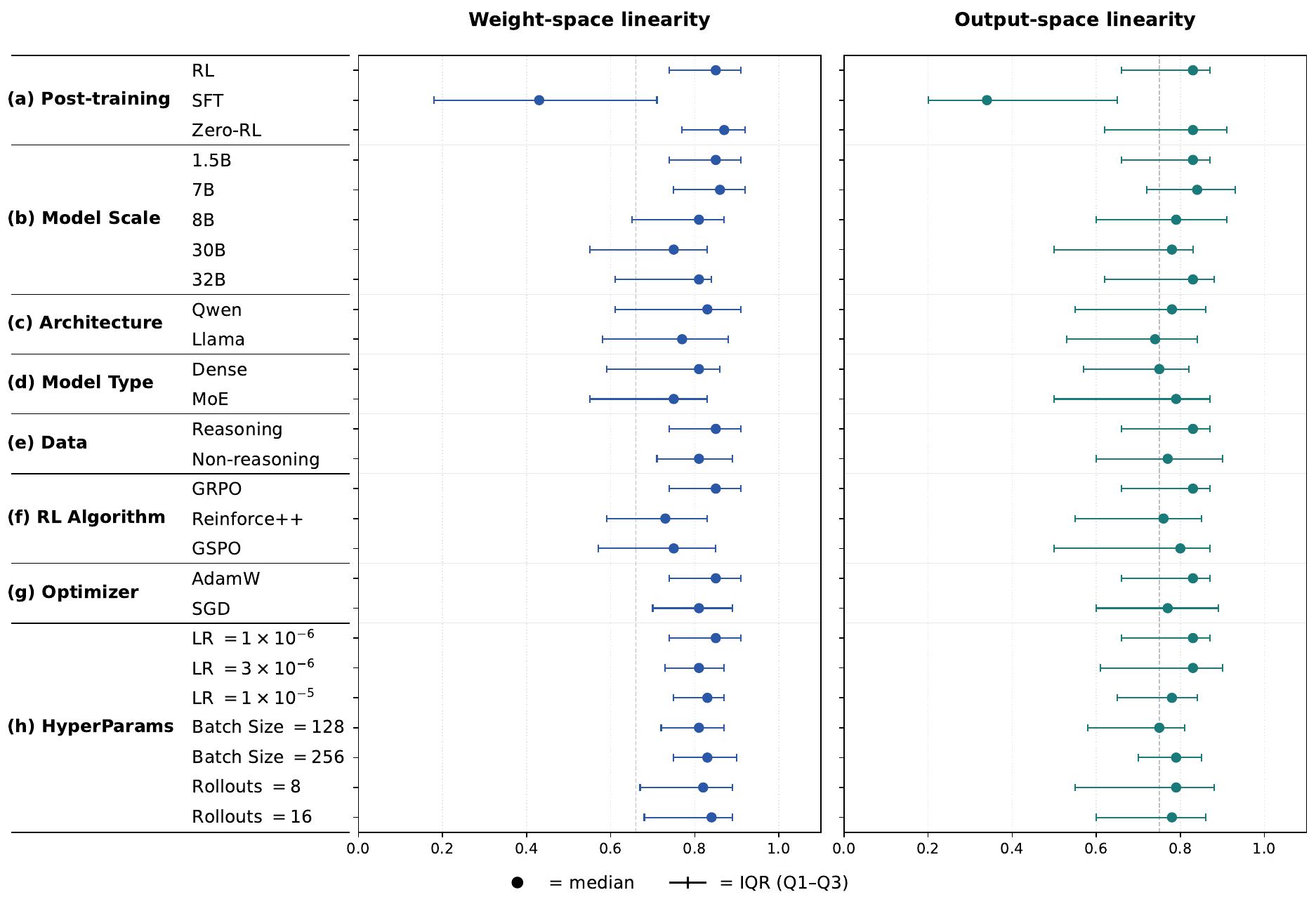}
    \caption{RLVR trajectory linearity across diverse experimental settings. Each row corresponds to a specific configuration grouped by category. \textbf{Left:} Weight-space linearity. \textbf{Right:} Output-space linearity. Horizontal lines represent the interquartile range (IQR) and dots indicate the median $R^2$ for each setting. \textbf{Vertical dashed lines} at $R^2 = 0.7$ highlight the threshold for strong linearity.}
    \vspace{-10pt}
    \label{fig:r2_robustness}
\end{figure}
We first investigate the linearity of model weights during RLVR. We collect intermediate checkpoints from the DeepScaleR pipeline~\citep{deepscaler2025} (Appendix~\ref{app:exp_setup}) and run a per-weight linear regression on the trainable parameters, using the coefficient of determination $R^2$ to quantify trajectory linearity.

As shown in Figure~\ref{fig:r2_robustness} (a, left), the distribution of $R^2$ values is concentrated near 0.8, with over 70\% of weights achieving $R^2>0.7$.
In contrast, the $R^2$ distribution for SFT on the same base model is centered primarily around 0.4, indicating significantly weaker linearity.
This surprising observation indicates that, for most parameters, the full training trajectory is well approximated by a linear trend.
A visualization of randomly sampled weights is provided in Figure~\ref{fig:weight_r2_case} (Appendix~\ref{app:weight_linearity}).

To further characterize linearity across different layers and components of LLMs, we compute the average $R^2$ for each layer.
As shown in Figure~\ref{fig:weight_r2_layer_heatmap} (Appendix~\ref{app:weight_linearity}), high linearity is consistent across the vast majority of model components, including attention and feed-forward (MLP) layers.

\subsection{Linearity in Output Logits and Intermediate Activations}
We next ask whether the linear evolution of weights translates to the linear updates of model outputs and intermediate activations.

To do so, we sample a fixed set of reasoning traces, generated by the base model, and use them as probes to track the evolution of log probabilities throughout training.
Details regarding probe construction and teacher-forced evaluation are provided in Appendix~\ref{app:output_linearity_details}.

As shown in Figure~\ref{fig:r2_robustness} (a, right), the distribution of token-level $R^2$ values is strongly concentrated near 0.8, indicating that output behavior evolves in a highly linear manner over the course of RLVR training.
Figure~\ref{fig:token_r2_case} presents examples of probing tokens: the log-probabilities of tokens such as ``wait'' and ``but'' increase steadily across RL steps, whereas those of ``earlier'' and ``alternatively'' decrease. These shifts may reflect changes in how the model structures and revises its reasoning.

We also observe that tokens with larger shifts in log-probabilities tend to display stronger linearity, suggesting that the most substantial changes in model output are often the most structurally organized.
A similar conclusion can be drawn on intermediate activations.
Due to page constraints, we defer discussion on activation linearity and the fine-grained analysis of different token categories to Appendix~\ref{app:output_linearity_details}.
Collectively, these results show that the linear regime of RLVR extends beyond weight space and is clearly reflected in token-level model behavior.

\subsection{Is the Observed Linearity a Coincidence?}
While we have observed strong linearity in RLVR, a critical question remains: could this phenomenon be a coincidence arising from a specific combination of base model, training data, RL algorithm and hyper-parameters? To address this, we verify the linearity of RLVR trajectories across a wide range of settings.
On the model side, we vary model scale (from 1.5B to 32B parameters), architecture (Qwen and Llama), and parameterization type (dense and MoE).
On the data side, we consider both reasoning and non-reasoning regimes, including datasets such as DeepScaleR and RLVR-IFeval.
On the algorithm side, we evaluate multiple mainstream RL methods, including GRPO, REINFORCE++, and GSPO.
Finally, on the training setup side, we experiment with different learning rates, batch sizes and numbers of rollouts.

As shown in Figure~\ref{fig:r2_robustness} (b)-(h), strong linearity persists across all evaluated configurations.
In every case, both weight-level and token-level $R^2$ remain above 0.7, demonstrating robustness to substantial variations in model scale, architecture, base model, training data, RL algorithm, and hyperparameters.
Detailed results for each setting are presented in Table~\ref{tab:r2_robustness} (Appendix~\ref{app:r2_robustness}), with further experimental and training details provided in Appendix~\ref{app:exp_setup}.

%% file: sections/4_origin.tex
\section{Origins of Linearity during RLVR}
\label{sec:origin}
Our preceding experiments establish pervasive linearity in model trajectories across various RLVR configurations.
Given the inherently nonlinear nature of Transformers, this phenomenon remains highly counterintuitive.
In this section, we uncover the core mechanisms underlying such linearity. 
We first eliminate multiple confounding factors that could account for weight-space linearity, then pinpoint a key origin of the linearity, and finally characterize how linear weight updates propagate to intermediate activations and final model outputs.

\subsection{Factors that Do NOT Induce RLVR Linearity}
\label{subsec:weight_origin}
We first evaluate three plausible candidate explanations for RLVR’s linear training dynamics: small weight update magnitudes, optimizer choice, and SFT initialization before reinforcement learning. Controlled experiments demonstrate that none of these factors serve as the fundamental cause.

\paragraph{Small weight update magnitudes.}
To stabilize training, RLVR typically employs a small learning rate (e.g., $\!1\mathrm{e}^{-6}$), which results in limited weight modification and could confine optimization to a locally linear region in parameter space.
However, we observe that even for weights exhibiting more than $5\%$ relative change, the average $R^2$ still exceeds $0.7$.
Furthermore, when increasing the learning rate to $1\mathrm{e}^{-5}$, training becomes less stable and performance degrades (e.g., AIME24 accuracy drops roughly $30\%$), yet the weight trajectories remain highly linear ($R^2 > 0.7$; see Figure~\ref{fig:r2_robustness} (h)).
These results indicate that the small magnitude of weight updates is not the primary cause of the linearity.

\paragraph{Optimizer choice.}
A second hypothesis is that AdamW’s momentum smooths noisy updates and reinforces consistent directions.
To isolate optimizer effects, we replace AdamW with vanilla SGD while keeping all other configurations fixed (Appendix~\ref{app:origin}).
The resulting $R^2$ distribution remains tightly concentrated above $0.7$ (Figure~\ref{fig:r2_robustness} (g)), so the linearity is not driven by AdamW.

\paragraph{SFT initialization prior to RL.}
A third hypothesis attributes linearity to SFT before RL, which may anchor the model in a flat and locally linear subspace.
We rule this out by running RLVR directly on the Qwen3-8B base model with no SFT initialization (denoted as ``zero-RL''; see Appendix~\ref{app:origin}). 
As Figure~\ref{fig:r2_robustness} (a) shows, the weight $R^2$ still predominantly lie above 0.7, so the linearity persists without any SFT initialization.

\subsection{Noisy Training Signal is a Key Reason for RLVR's Linearity}\label{subsec:noisy_signals}
Beyond the factors above, another key property of RLVR is that its training signal is highly abstract and noisy. 
Leading RLVR algorithms, such as GRPO and REINFORCE++, assign token-level credit solely based on final answer correctness. 
This design introduces high variance: locally correct token choices may receive negative rewards when subsequent errors lead to an incorrect final answer, while locally incorrect ones may receive undeserved positive rewards when later reflection steps recover the outcome. 
In this part, we give an intuitive and formal explanation for why the noisy training signals of RLVR can potentially lead to its linearity. 
Then, we establish the empirical relationship between the noise level of training signal and the linearity of the training trajectories.

Under noisy supervision, models prioritize \emph{frequent, high-level patterns} and act as low-pass filters that suppress sporadic, fine-grained variations~\citep{han2025on, pmlr-v70-arpit17a}. This aligns with recent work on credit assignment, which shows that uniformly distributing the final outcome reward across tokens prevents models from learning rare, low-level actions~\citep{parthasarathi2025grpolambdacreditassignmentimproves, li2026outcomegroundedadvantagereshapingfinegrained}. Concretely, the per-step gradient is highly noisy at the token level, but, when averaged, individual sample gradients largely cancel and the expected update concentrates in a low-dimensional subspace spanned by the most frequent successful patterns. Within that subspace, the optimization direction is approximately fixed, which produces a stable linear drift in weight space.

\paragraph{Formal analysis.}
We make the above intuition rigorous via a lazy-training analysis of GRPO. Let $\varphi_{\theta_0}(\tau) := \nabla_\theta \log \pi_\theta(\tau)\big|_{\theta=\theta_0}$, $v_R = \mathbb{E}_{\pi_{\theta_0}}[\varphi_{\theta_0}(\tau) \mid R(\tau)=1]$, and $\delta = \theta - \theta_0$. Under a standard NTK-style lazy assumption~\citep{jacot2018neural, chizat2019lazy, malladi2023kernel}, we show that the GRPO gradient remains close to the fixed direction $\frac{\bar{p}_\theta}{\sigma_\theta} v_R$:
\begin{theorem}\label{main_theorem}
Under a standard NTK-style lazy assumption, let $v_R = \mathbb{E}_{\pi_{\theta_0}}[\varphi_{\theta_0}(\tau) \mid R(\tau)=1]$ and $K_S = \text{Cov}_{\pi_{\theta_0}}(\varphi_{\theta_0} \mid R(\tau)=1)$. The GRPO gradient satisfies:
\begin{equation}
    \| g_{\text{GRPO}}(\theta) - \frac{\bar{p}_\theta}{\sigma_\theta} v_R \| \le \frac{1}{\sigma_\theta} \left(\|K_S\|_{\text{op}} + 2 \kappa \right)\|\delta\| + \frac{2 C L }{\sigma_\theta} \|\delta\|^2.
\end{equation}
where $C > 0$ is a constant capturing the second-order curvature of the policy distribution.
\end{theorem}
The full assumption and proof are deferred to Appendix~\ref{append:theorem}. Two consequences follow. First, during training, the parameter updates are dominated by the \emph{fixed} vector $v_R$, i.e.\ the average gradient of successful responses at initialization. Second, the scalar $\frac{\bar{p}_\theta}{\sigma_\theta} = \sqrt{\frac{\bar{p}_\theta}{(1-\bar{p}_\theta)}}$ stays in a moderate range (e.g.\ $0.5$--$2.0$ for $\bar{p}_\theta\in[0.2,0.8]$), so the gradient magnitude does not fluctuate drastically. Together, $\theta$ evolves in a nearly linear fashion away from $\theta_0$, justifying the observed linear regime. 

\paragraph{Empirical verification.}
\begin{wrapfigure}{r}{0.5\textwidth}
    \centering
    \includegraphics[width=\linewidth]{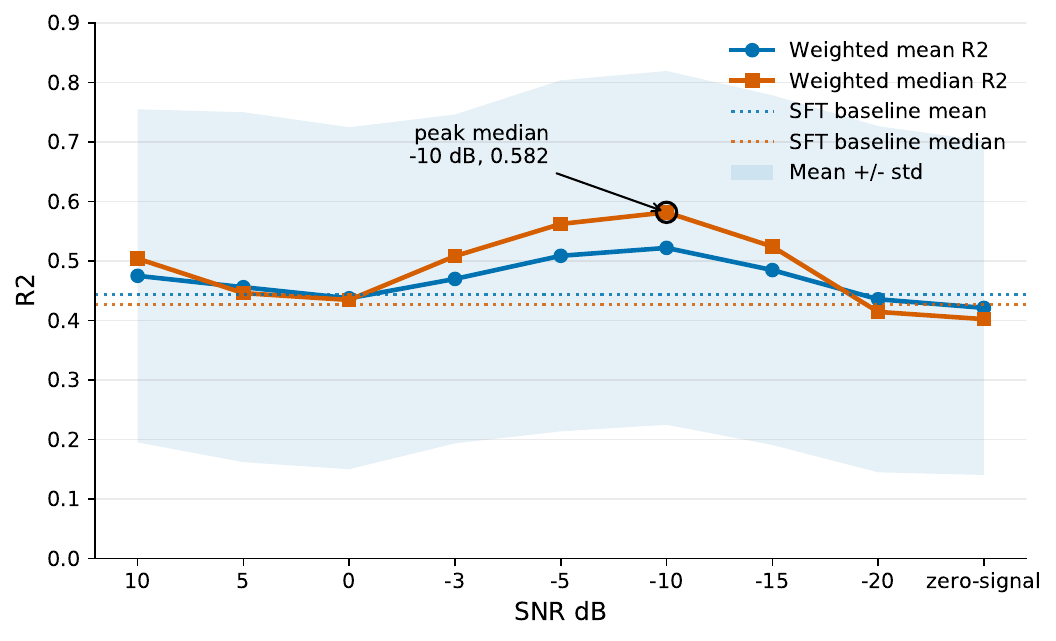}
    \caption[Token-level SNR controls weight-space linearity.]{\textbf{Token-level SNR controls weight-space linearity.} The active-parameter weighted mean and weighted median $R^2$ are shown across token-level random-sign SNR settings. The shaded band denotes the mean $\pm$ standard deviation, and dotted horizontal lines indicate the vanilla SFT baseline.}
    \label{fig:sft+noise_snr_r2}
\end{wrapfigure}
To validate this account, we test whether injecting reward-style noise into a normally-deterministic objective increases trajectory linearity.
Reducing the noise of RLVR via per-token Monte Carlo Tree Search would be the most direct test but is computationally infeasible because of the explosion in rollouts.
We therefore go in the other direction: starting from Supervised Fine-Tuning (SFT), whose gradients are highly deterministic and far less noisy than RLVR, we artificially inject controlled noise into the SFT signal (Appendix~\ref{app:origin}).
Token-level and sequence-level noise increase the median $R^2$ of weights from a baseline of 0.426 to 0.582 and 0.568, respectively. Furthermore, as illustrated in Figure~\ref{fig:sft+noise_snr_r2}, $R^2$ exhibits a non-monotonic trend: it initially rises to a peak as noise increases before eventually declining.
This direct evidence confirms that gradient noise actively contributes to weight linearity by filtering out fine-grained optimization directions, locking the model into amplifying a robust, low-dimensional drift, exactly as our analysis predicts.

\subsection{How Does Weight Linearity Lead to Output and Activation Linearity?}
Even if model weights update linearly, it is still surprising that intermediate activations and final outputs evolve linearly during RLVR, given the strongly nonlinear computation of Transformers.
For analysis, we pick a linear layer $y_t = W_t x_t$ in an MLP block as an example, where $x_t$, $y_t$, and $W_t$ are the input, output, and weight matrix at step $t$. Even though both $W_t = W_0 + \Delta W \cdot t$ and $x_t = x_0 + \Delta x \cdot t$ are linear in $t$, the output expands as
\begin{equation}
y_t = (W_0 + \Delta W \cdot t)(x_0 + \Delta x \cdot t)
= W_0 x_0 + (\Delta W \, x_0 + W_0 \, \Delta x) \, t + (\Delta W \, \Delta x) \, t^2
\end{equation}
which is quadratic rather than linear in $t$.

In practice, however, we find that the quadratic term $\Delta W \Delta x \cdot t^2$ is negligibly small compared to the 
\begin{wrapfigure}{r}{0.5\textwidth}
    \centering
    \vspace{-1em}
    \includegraphics[width=\linewidth]{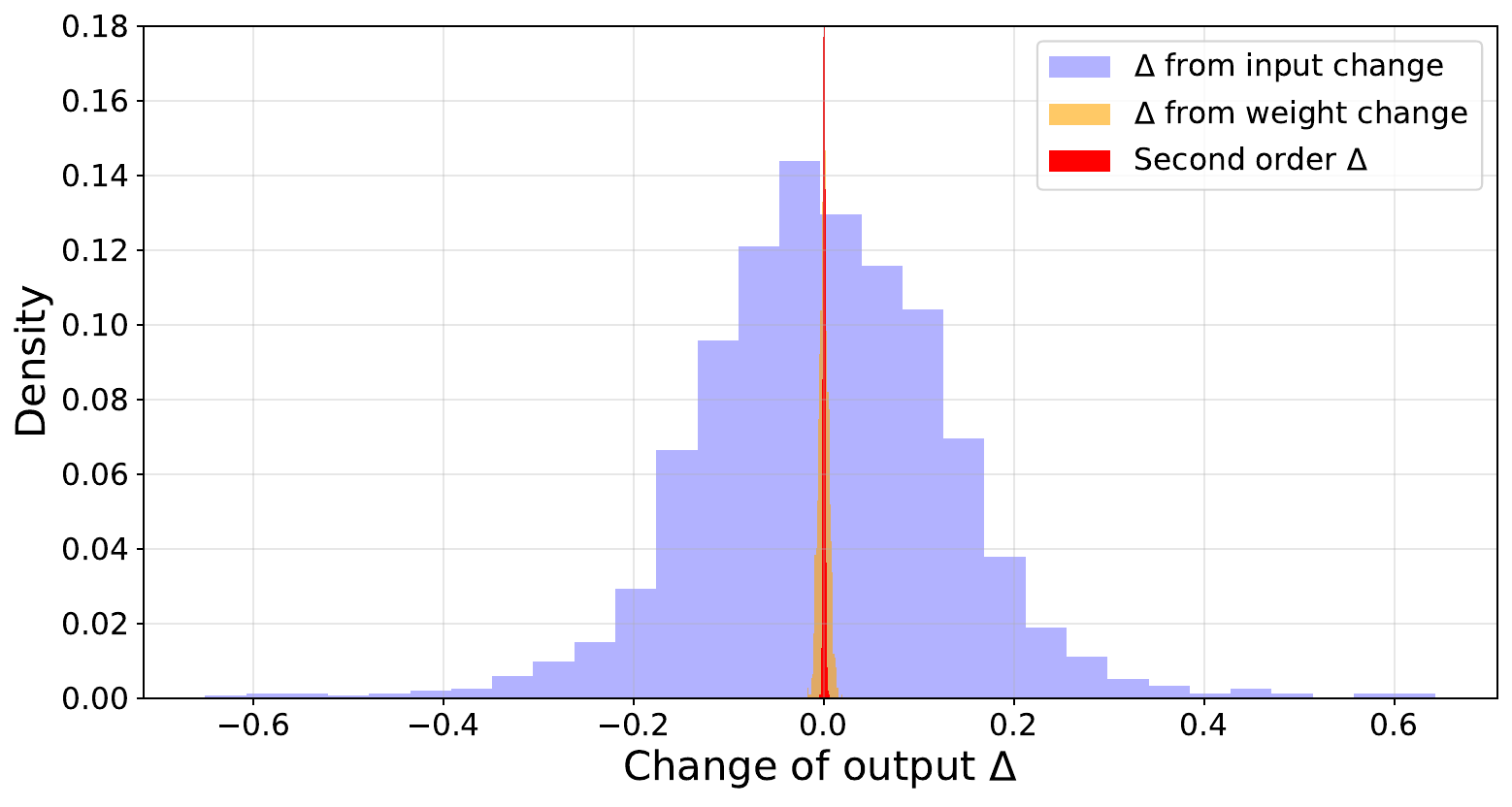}
    \caption{The source of output changes in a representative LLM layer.}
    \label{fig:origin_output_linearity}
    \vspace{-1em}
\end{wrapfigure}
linear term $W_0 \Delta x \cdot t$, which dominates the change of the output $y$. Figure~\ref{fig:origin_output_linearity} illustrates the contribution of the first- and second-order terms to the output change of a linear layer in the transformer.
We can see that the output change is driven primarily by the first-order impact of input and weight changes, while the second-order term remains uniformly small across samples.

We can also see that the change in output $y$ mainly results from the change in the input $x$, which accumulates small weight adjustments from previous layers.
The exact same derivation applies to the attention and embedding layers.

%% file: sections/5_extrapolation.tex
\section{Predictive Extrapolation of RLVR Trajectories}
\label{sec:extrapolation}
To establish that the observed linear structure is a fundamental property of RLVR rather than a retrospective artifact, we rigorously test its predictive power.
If the linear regime is mechanistically grounded, it should be possible to forecast future model states directly from earlier trajectory segments.
In this section, we test the predictive capability in both output space and weight space.
To ensure optimal RL performance, we utilize intermediate checkpoints from DeepScaleR~\citep{deepscaler2025} as our \textit{Standard RL} baseline, upon which all trajectory extrapolations are anchored.
Specifically, we check whether we can predict the output logits and weights in a future step, without actual training.
We evaluate our extrapolation methods on AIME24/25~\citep{aime24, aime25}, MATH500~\citep{math500} and LiveCodeBench~\citep{livecodebench} benchmarks. Full hyperparameter configurations and evaluation details for these runs are deferred to Appendix~\ref{app:exp_setup}.

\subsection{Output-space Extrapolation}
\begin{wraptable}{r}{0.47\textwidth}
  \centering
  \vspace{-1.5em}
  \small
  \caption{Output-space extrapolation.}
  \setlength{\tabcolsep}{3pt}
  \begin{tabular}{l r r r}
  \toprule
  \multicolumn{1}{c}{\textbf{Benchmark}}
    & \multicolumn{1}{c}{\textbf{Std.\ RL}}
    & \multicolumn{1}{c}{\textbf{Ours}}
    & \multicolumn{1}{c}{\textbf{\%Imp}} \\
  \midrule
  AIME24 (avg@64) & 0.4193 & 0.4458 & \textbf{$+$6.3\%} \\
  AIME25 (avg@64) & 0.3135 & 0.3333 & \textbf{$+$6.3\%} \\
  MATH500 (avg@64)& 0.8803 & 0.8900 & \textbf{$+$1.1\%} \\
  LCB\,(avg@4)    & 0.1786 & 0.1976 & \textbf{$+$10.6\%} \\
  \midrule
  Average         & 0.4479 & 0.4667 & \textbf{$+$4.2\%} \\
  \bottomrule
  \end{tabular}
  \vspace{-0.7em}
  \label{tab:extrapolation}
\end{wraptable}
We start by predicting token logits $l_{t'}$ at a future time step $t'$ based on logits $l_{t_0}$ and $l_{t_1}$ at existing checkpoints at time $t_0$ and $t_1$:
\begin{equation}
    \mathbf{l}_{t'} = \mathbf{l}_{t_0} + \alpha (\mathbf{l}_{t_1} - \mathbf{l}_{t_0}).
\end{equation}
Here $\alpha = \frac{t' - t_0}{t_1 - t_0} > 1$ is the extrapolation ratio.

This formulation provides a simple, training-free mechanism to predict the policy at a future stage.

Table~\ref{tab:extrapolation} compares the extrapolated model with the model from real training on AIME24/25, MATH500, and LiveCodeBench. Across all tasks, the extrapolated model consistently outperforms the standard RL training. This not only confirms the predictive power of output linearity but also offers a path to overcome the instability of late-stage RL training to further boost reasoning performance.

\subsection{Weight-Space Extrapolation}
\label{sec:weight_extra}
A similar extrapolation experiment is conducted in weight space.
For a future time step $t'$, we directly predict model weights at $t'$, from two previous checkpoints at time steps $t_0 < t_1$:
\begin{equation}
    \mathbf{W}_{t'} = \mathbf{W}_{t_0} + \beta (\mathbf{W}_{t_1} - \mathbf{W}_{t_0}),
\end{equation}
where $\beta = \frac{t' - t_0}{t_1 - t_0} > 1$ is the extrapolation ratio in weight space. Compared with output-space extrapolation, the extrapolated parameters $\mathbf{W}_{t'}$ instantiate a complete, fully functional model ready for downstream operations like training and deployment.

\begin{wrapfigure}{r}{0.51\textwidth}
  \centering
  \vspace{-1em}
  \includegraphics[width=\linewidth]{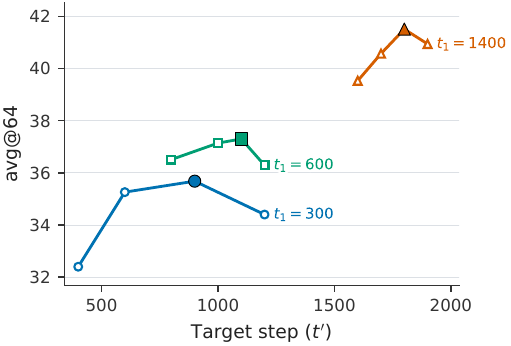}
  \caption{Weight-space extrapolation performance on AIME24 across different target steps. Each curve corresponds to a different choice of \(t_1\), and the filled marker denotes the peak avg@64 on that curve.
}
  \label{fig:extrapolation}
  \vspace{-2em}
\end{wrapfigure}
Figure~\ref{fig:extrapolation} demonstrates the performance of this lookahead model on AIME24 as the target step $t'$ varies.
A clear inverted-U pattern emerges: moderate extrapolation strictly improves performance, confirming the predictive validity of the linear direction. 
However, pushing the projection too far eventually causes degradation (e.g., performance peaks around $t'=900$ for anchors at $t_0=0, t_1=300$).
This is because the estimation of weight slopes is not perfectly accurate in limited time steps $t_1-t_0$. As a result, when the extrapolation ratio $\beta$ gets high, errors are also amplified in the prediction of $\mathbf{W}_t$. 
Detailed empirical guidelines for selecting the extrapolation hyperparameters (i.e., $t_0$, $t_1$, and $\beta$) under a given training budget are provided in Appendix~\ref{app:weight_extra}.

\subsection{Periodic Re-grounding}\label{sec:periodic-re-grounding}
To mitigate performance degradation from amplified extrapolation errors, we introduce a dynamic approach that interleaves gradient-free weight extrapolation within standard RL optimization. 
Formally, this process operates in cycles of period $C=m+n$ (configured in Appendix~\ref{app:regrounding}). Each cycle begins with $m$ steps of standard gradient descent to align with the true reward signal, followed by $n$ steps of linear projection to accelerate progress and reduce computational overhead. Let $k$ denote the current global step and $\eta$ the learning rate. The update rule for the model parameters $\mathbf{W}_{k+1}$ using the RL objective function $\mathcal{L}_{\text{RL}}$ is defined as:
\begin{equation}
\mathbf{W}_{k+1} =
\begin{cases}
\mathbf{W}_k - \eta \nabla \mathcal{L}_{\text{RL}}(\mathbf{W}_k), & \text{if } (k \bmod C) < m, \\[1em]
\mathbf{W}_k + \frac{1}{m} \left( \mathbf{W}_{\tau+m} - \mathbf{W}_{\tau} \right), & \text{otherwise},
\end{cases}
\label{eq:update_rule}
\end{equation}

\begin{wrapfigure}{r}{0.5\textwidth}
    \centering
    \vspace{-1em}
    \includegraphics[width=\linewidth]{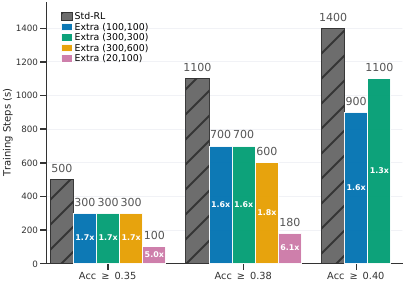}
    \caption{\textbf{Effect of periodic re-grounding on RL training efficiency.} Bars are grouped by target AIME24 accuracy levels ($\geq$0.35, $\geq$0.38, $\geq$0.40). Annotated values indicate the speedup factor, defined as the ratio of steps required under standard RL to those with periodic re-grounding.}
    \label{fig:regrounding_efficiency}
\end{wrapfigure}

\begin{table}[t]
\centering
\caption{\textbf{Performance under Fixed Training Budgets.} We evaluate our proposed approach (extrapolation via periodic re-grounding) against the standard RL baseline. Under fixed gradient computation budgets (actual training steps $s$), our method consistently yields a higher average performance than the standard optimization path.}
\vspace{8pt}
\resizebox{\textwidth}{!}{
\begin{tabular}{l c l c c c c c c}
\toprule
\textbf{Phase} & \textbf{Steps} & \textbf{Method}
  & \textbf{AIME24} & \textbf{AIME25} & \textbf{MATH500} & \textbf{LCB}
  & \textbf{Avg} & \textbf{\%Imp} \\
 & ($s$) &
  & \textbf{(Avg@64)} & \textbf{(Avg@64)} & \textbf{(Avg@64)} & \textbf{(Avg@4)}
  & \textbf{Score} & \\
\midrule
\multirow{2}{*}{Early} & \multirow{2}{*}{200}
  & Standard RL & 0.3172 & 0.2536 & 0.8421 & 0.2714 & 0.4211 & — \\
  & & \cellcolor{extrap}Ours & \cellcolor{extrap}0.3318 & \cellcolor{extrap}0.2979 & \cellcolor{extrap}0.8611 & \cellcolor{extrap}0.2619 & \cellcolor{extrap}0.4382 & \cellcolor{extrap}+5.2\% \\
\midrule
\multirow{2}{*}{Mid} & \multirow{2}{*}{400}
  & Standard RL & 0.3391 & 0.2682 & 0.8525 & 0.2810 & 0.4352 & — \\
  & & \cellcolor{extrap}Ours & \cellcolor{extrap}0.3672 & \cellcolor{extrap}0.3005 & \cellcolor{extrap}0.8658 & \cellcolor{extrap}0.2905 & \cellcolor{extrap}0.4560 & \cellcolor{extrap}+6.3\% \\
\midrule
\multirow{2}{*}{Late} & \multirow{2}{*}{800}
  & Standard RL & 0.3490 & 0.2932 & 0.8664 & 0.2821 & 0.4477 & — \\
  & & \cellcolor{extrap}Ours & \cellcolor{extrap}0.3984 & \cellcolor{extrap}0.3094 & \cellcolor{extrap}0.8667 & \cellcolor{extrap}0.2905 & \cellcolor{extrap}0.4663 & \cellcolor{extrap}+5.7\% \\
\midrule
\multirow{2}{*}{Converged} & \multirow{2}{*}{1200}
  & Standard RL & 0.3828 & 0.2995 & 0.8658 & 0.2857 & 0.4585 & — \\
  & & \cellcolor{extrap}Ours & \cellcolor{extrap}0.4120 & \cellcolor{extrap}0.3120 & \cellcolor{extrap}0.8731 & \cellcolor{extrap}0.2762 & \cellcolor{extrap}0.4683 & \cellcolor{extrap}+2.3\% \\
\bottomrule
\end{tabular}
}
\vspace{-10pt}
\label{tab:regrounding_performance}
\end{table}
Table~\ref{tab:regrounding_performance} provides compelling validation for this trajectory-level regularity, demonstrating substantial improvements in training efficiency. Under matched training budgets, periodic re-grounding consistently outperforms standard RL on all three math benchmarks~(AIME24, AIME25, and MATH500) while maintaining a similar level of performance on LiveCodeBench. Figure~\ref{fig:regrounding_efficiency} further 
illustrates strong speedups across different re-grounding schedules. For example, to match the standard RL's AIME24 accuracy of 0.40, a $(100,100)$ schedule requires only 900 actual RL steps---a $1.6\times$ reduction in gradient-based training. Even a more aggressive $(20,100)$ schedule reaches $>0.38$ AIME24 accuracy with just 180 RL steps, matching the baseline trained for 1100 steps and yielding a $6.1\times$ speedup.

Taken together, these extrapolative experiments offer a clarified view of RLVR dynamics. 
Output-space extrapolation provides a lightweight function-space mechanism that effectively bypasses late-stage model collapse while yielding practical performance improvements.
Meanwhile, weight-space extrapolation more directly tracks the underlying optimization trajectory, rapidly producing a lookahead model that closely matches the performance of further GPU-intensive RL training. By repeatedly projecting extrapolated weights back to normal RL trajectories to address long-horizon inaccuracies, periodic re-grounding maintains this alignment and significantly accelerates the overall training process over extended runs.

%% file: sections/6_conclusion.tex
\section{Conclusion}
In this paper, we presented a comprehensive trajectory-level analysis of RLVR.
We identified a pervasive \textit{robust linear regime} wherein both model parameters and output space evolve linearly over the course of training.
We support this observation with both theoretical derivation and empirical evidence, demonstrating that the noisy reward signals in RLVR effectively regularize optimization by filtering out local fluctuations and maintaining a stable macroscopic drift.
Leveraging this foundational insight, we introduced practical algorithms for predictive extrapolation in both weight and output spaces. Specifically, weight-space extrapolation, integrated with a periodic re-grounding, can match RL performance with lower overhead, while output-space extrapolation provides a lightweight intervention that prevents late-stage model collapse, allowing models to consistently outperform standard RL.
We provide a deeper discussion of the broader implications of our findings in Appendix~\ref{sec:implication}. We also discuss the potential limitations of this work in Appendix~\ref{sec:limitation}.

%% file: sections/7_appendix.tex
\section{Experimental Setup}\label{app:exp_setup}
We follow experimental setup in DAPO~\citep{NEURIPS2025_a4277440}, including the benchmarks and evaluation setups.

\subsection{Models}
In our empirical evaluation, we utilize a diverse set of base models to ensure the broad applicability of our findings. The evaluated models vary significantly across three main dimensions:
\begin{itemize}
    \item \textbf{Varying Parameter Scales}: Our experiments cover models from lightweight sizes (1.5B) to larger scales (up to 32B total parameters), allowing us to analyze the scaling behavior of our observation.
    \item \textbf{Diverse Architectures}: To ensure our conclusions are not architecture-specific, we select models originating from different foundational families, including the Qwen/Qwen3 series and the LLaMA architecture. We also include NVIDIA Nemotron reasoning models built on top of Qwen backbones to test whether the observed phenomenon persists under different post-training pipelines and training recipes.
    \item \textbf{Dense vs. Mixture-of-Experts (MoE)}: While the majority of our evaluated models are dense networks, we specifically include Qwen3-30B-A3B—an MoE architecture with 30 billion total parameters and approximately 3 billion active parameters per token—to validate our observation on sparse models.
\end{itemize}

\begin{table}[htbp]
    \centering
    \caption[Base models evaluated in our study.]{\textbf{Base models evaluated in our study.} The table outlines the base models used to evaluate RLVR trajectory linearity. For MoE models (e.g., Qwen3-30B-A3B), we report both the total parameter count and the number of active parameters activated during inference.}
    \label{tab:evaluated_models}
    \begin{tabular}{llcc}
        \toprule
        \textbf{Model} & \textbf{Architecture} & \textbf{Params (Total / Active)} & \textbf{Type} \\
        \midrule
        \texttt{DeepSeek-R1-Distill-Qwen-1.5B}  & Qwen     & 1.5B / 1.5B     & Dense \\
        \texttt{Open-Nemotron-1.5B}             & Nemotron & 1.5B / 1.5B     & Dense \\
        \texttt{DeepSeek-R1-Distill-Qwen-7B}    & Qwen     & 7B / 7B         & Dense \\
        \texttt{DeepSeek-R1-Distill-Llama-8B}   & LLaMA    & 8B / 8B         & Dense \\
        \texttt{Qwen3-8B}                       & Qwen3    & 8B / 8B         & Dense \\
        \texttt{Qwen2.5-32B}                    & Qwen     & 32B / 32B       & Dense \\
        \texttt{Qwen3-30B-A3B}                  & Qwen3    & 30B / $\sim$3B  & MoE   \\
        \bottomrule
    \end{tabular}
\end{table}

\subsection{RL algorithms}
To ensure that the generalizability of our observations is not an artifact of a specific optimization objective or training dynamic, we conduct comprehensive experiments across a diverse spectrum of RL algorithms. The evaluated frameworks include:
\begin{itemize}
    \item \textbf{GRPO}~\citep{shao2024deepseekmathpushinglimitsmathematical}: Representing the state-of-the-art in reasoning-focused LLM alignment, GRPO optimizes the policy by normalizing rewards across a sampled group of generated outputs, enabling effective updates without relying on a value model.
    \item \textbf{REINFORCE++}~\citep{hu2025reinforcestabilizingcriticfreepolicy}: This framework proposes Global Advantage Normalization to replace the local group normalization used in GRPO. It effectively corrects the bias introduced by per-prompt normalization in existing critic-free approaches while maintaining computational efficiency by eliminating the critic model.
    \item \textbf{GSPO}~\citep{zheng2025groupsequencepolicyoptimization}: Designed specifically to enhance the training stability of Mixture-of-Experts (MoE) models, GSPO elevates the optimization granularity from the token level to the sequence level, ensuring more robust convergence.
\end{itemize}

\subsection{Training Details}
The training hyperparameters are shown in the Table~\ref{tab:hyperparams}. We adapt our training codebase from VERL and follow the training recipe of three different RL algorithms, including GRPO, GSPO, and REINFORCE++.

Specifically, we collect the intermediate checkpoints from DeepScaleR~\citep{deepscaler2025}, JustRL~\citep{he2025justrlscaling15bllm} and Skywork-OR1-7B~\citep{he2025skyworkopenreasoner1} to ensure optimal performance.

\begin{table}[t]
\centering
\caption[Hyperparameter settings for RL training.]{\textbf{Hyperparameter settings.} These settings are applied consistently across GRPO, REINFORCE++, and GSPO.}
\begin{tabular}{lc}
\toprule
\textbf{Hyperparameter} & \textbf{Value} \\
\midrule
KL Loss & No \\
Entropy Regularization & No \\
Global Batch Size & [128, 256] \\
PPO Mini-batch Size & 64 \\
Max Response Length & 16K \\
Learning Rate & $1 \times 10^{-6}$ (Constant) \\
Clip Ratio Range & $[0.8, 1.28]$ \\
Temperature & 1.0 \\
Rollout ($N$) & [5, 8, 16] \\
\bottomrule
\end{tabular}
\label{tab:hyperparams}
\end{table}

\subsection{Computational Resources}
\label{app:compute_resources}
Table~\ref{tab:compute_resources} summarizes the computational resources used for the main training experiment groups. Unless otherwise specified, each GPU run used 8 NVIDIA H200 GPUs with 141GB memory per GPU. We report approximate GPU-hours because wall-clock time varied with cluster scheduling and evaluation batch composition. We report only the compute incurred by our post-training and re-grounding runs.

\begin{table}[t]
\centering
\footnotesize
\setlength{\tabcolsep}{3pt}
\caption{\textbf{Computational resources.} Costs are approximate GPU-hours. ``Cost basis'' gives the per-run or per-group accounting used to estimate the aggregate compute for the reported experiments.}
\label{tab:compute_resources}
\begin{tabular}{@{}p{0.22\textwidth}p{0.43\textwidth}p{0.17\textwidth}p{0.11\textwidth}@{}}
\toprule
\textbf{Experiment group} & \textbf{Scope} & \textbf{Cost basis} & \textbf{Total cost} \\
\midrule
Post-Training runs & All runs reported in Table~\ref{tab:r2_robustness} & $\sim$2,000/run $\times$ 13 & $\sim$26,000 \\
Weight-space extrapolation & Re-grounding RL schedules reported in Table~\ref{tab:regrounding_performance} & $\sim$700/run $\times$ 4 & $\sim$2,800 \\
\bottomrule
\end{tabular}
\end{table}

\subsection{Existing Assets and Licenses}
\label{app:assets_licenses}
We use existing models, checkpoints, datasets, benchmarks, and code under their public release terms, and cite the corresponding original sources throughout the paper. Table~\ref{tab:asset_licenses} summarizes the main external assets used in our experiments.

\begin{table}[t]
\centering
\scriptsize
\setlength{\tabcolsep}{3pt}
\caption{\textbf{Existing assets and licenses.} We use these assets for research evaluation and training-dynamics analysis only, without redistributing third-party model weights or benchmark data beyond their upstream release terms.}
\label{tab:asset_licenses}
\begin{tabular}{p{0.22\textwidth}p{0.42\textwidth}p{0.28\textwidth}}
\toprule
\textbf{Asset group / use} & \textbf{Assets / sources} & \textbf{License / terms} \\
\midrule
Base models for trajectory analysis and evaluation & DeepSeek-R1-Distill models~\citep{Guo_2025}; Qwen/Qwen3/Qwen2.5 models~\citep{yang2025qwen3technicalreport}; Open-Nemotron/OpenMath-Nemotron models~\citep{moshkov2025aimo2winningsolutionbuilding} & MIT for DeepSeek-R1 distills, with upstream Qwen Apache-2.0 and Llama license terms where applicable; Apache-2.0 for Qwen releases; CC-BY-4.0 with Apache-2.0 information for Open-Nemotron/OpenMath-Nemotron. \\
RL checkpoints and training data & DeepScaleR~\citep{deepscaler2025}, JustRL~\citep{he2025justrlscaling15bllm}, Skywork-OR1~\citep{he2025skyworkopenreasoner1}, DAPO-Math-17k~\citep{NEURIPS2025_a4277440}, and RLVR-IFeval~\citep{lambert2024tulu} & MIT for DeepScaleR; Apache-2.0 for JustRL, Skywork-OR1 code releases, and DAPO-Math-17k; ODC-BY for RLVR-IFeval. \\
Evaluation benchmarks & AIME24/25~\citep{aime24,aime25}, MATH500~\citep{math500}, and LiveCodeBench~\citep{livecodebench} & Public benchmark use terms for AIME24/25; MIT for MATH/MATH500 source releases; MIT for LiveCodeBench. \\
Training codebase & VERL (\url{https://github.com/volcengine/verl}) & Apache-2.0. \\
\bottomrule
\end{tabular}
\end{table}

\subsection{Evaluation Benchmarks}
To rigorously assess our proposed extrapolation techniques, we conducted an evaluation across a diverse suite of benchmarks. Importantly, the primary objective of this extrapolation is to verify that the linearity of the model's performance scaling is predictable, and to demonstrate that this predictability can be effectively leveraged to accelerate RLVR training. The specific focus and rationale for each benchmark category are outlined below:
\begin{itemize}
    \item \textbf{Math Reasoning}: AIME24/25~\citep{aime24, aime25} consist of competition-level mathematics problems from the American Invitational Mathematics Examination, which require advanced, multi-step mathematical problem-solving; and MATH500~\citep{math500} is a curated, representative subset of 500 problems from the MATH dataset, spanning various mathematical disciplines and difficulty levels.
    \item \textbf{Code Generation}: LiveCodeBench (v5, Oct 2024 – Feb 2025)~\citep{livecodebench} is a continuously updated benchmark for code generation based on real-world competitive programming platforms (e.g., LeetCode, Codeforces).
\end{itemize}

\subsection{Evaluation Setup}
\label{app:eval_setup}
We evaluate models on four standard mathematical and code reasoning benchmarks commonly used for assessing reasoning capabilities: AIME'24, AIME'25, MATH500, and LiveCodeBench. All evaluations are conducted in a zero-shot setting. For each question, the maximum generation length is set to $32,768$ tokens under a temperature of $0.6$, a top-p value of $0.95$.

We report Avg@$k$ and Pass@$k$, defined as follows: Pass@$k$ measures the proportion of problems where at least one correct solution exists among the top-$k$ samples, reflecting the model's potential coverage. Avg@$k$ denotes the average accuracy (expected Pass@1) calculated over the $k$ samples, reflecting the model's stability.

\section{Additional Results on Empirical Phenomenon}\label{app:empirical_law}
\subsection{Linearity in Weights}
\label{app:weight_linearity}
\begin{figure}[t]
    \centering
    \includegraphics[width=\columnwidth]{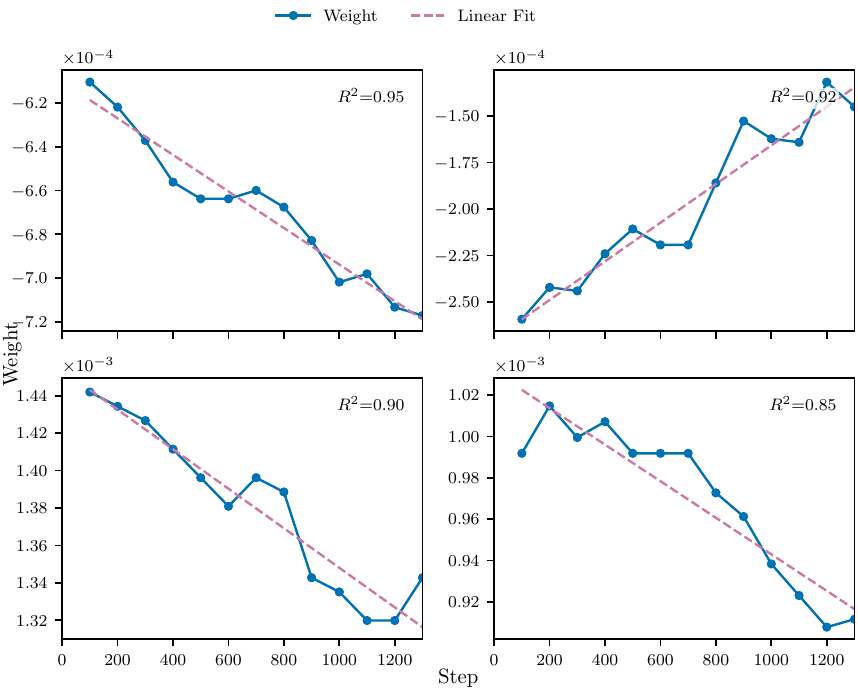}
    \caption[Dynamics of high-linearity weight dimensions.]{\textbf{Dynamics of high-linearity weight dimensions.} Representative weight trajectories are shown across training steps. Solid blue curves plot the exact weight values, dashed pink lines represent the corresponding linear fits, and $R^2$ values indicate the goodness of fit.}
    \label{fig:weight_r2_case}
\end{figure}

\begin{figure}[t]
    \centering
    \includegraphics[width=\columnwidth]{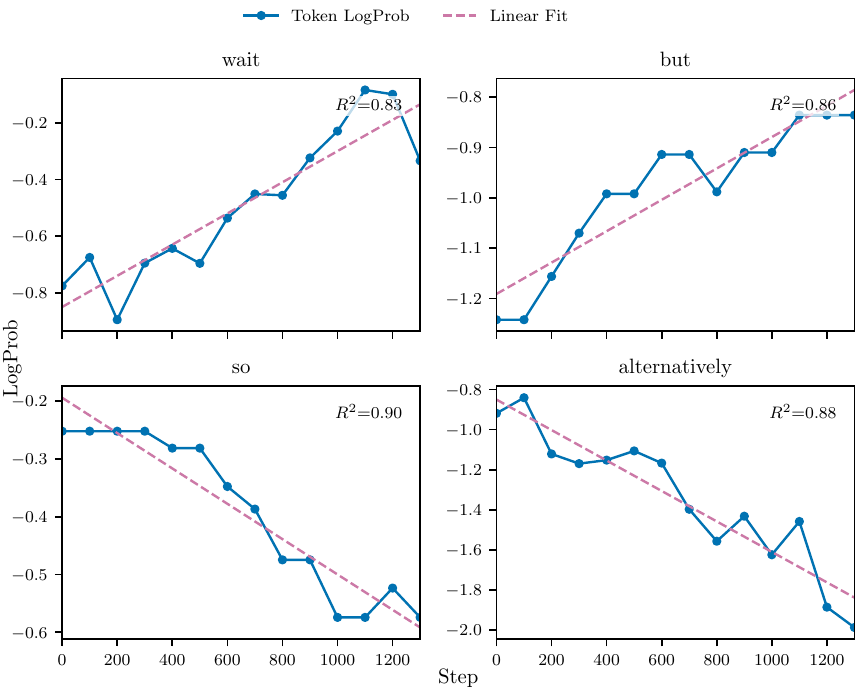}
    \caption[Representative high-linearity token log probabilities.]{\textbf{Representative high-linearity token log probabilities.} Shown are the log-probability dynamics of selected tokens across training steps. Solid blue curves plot the actual values, dashed pink lines represent their linear fits, and $R^2$ values indicate the goodness of fit.}
    \label{fig:token_r2_case}
\end{figure}
\begin{figure}[t]
    \centering
    \includegraphics[width=\columnwidth]{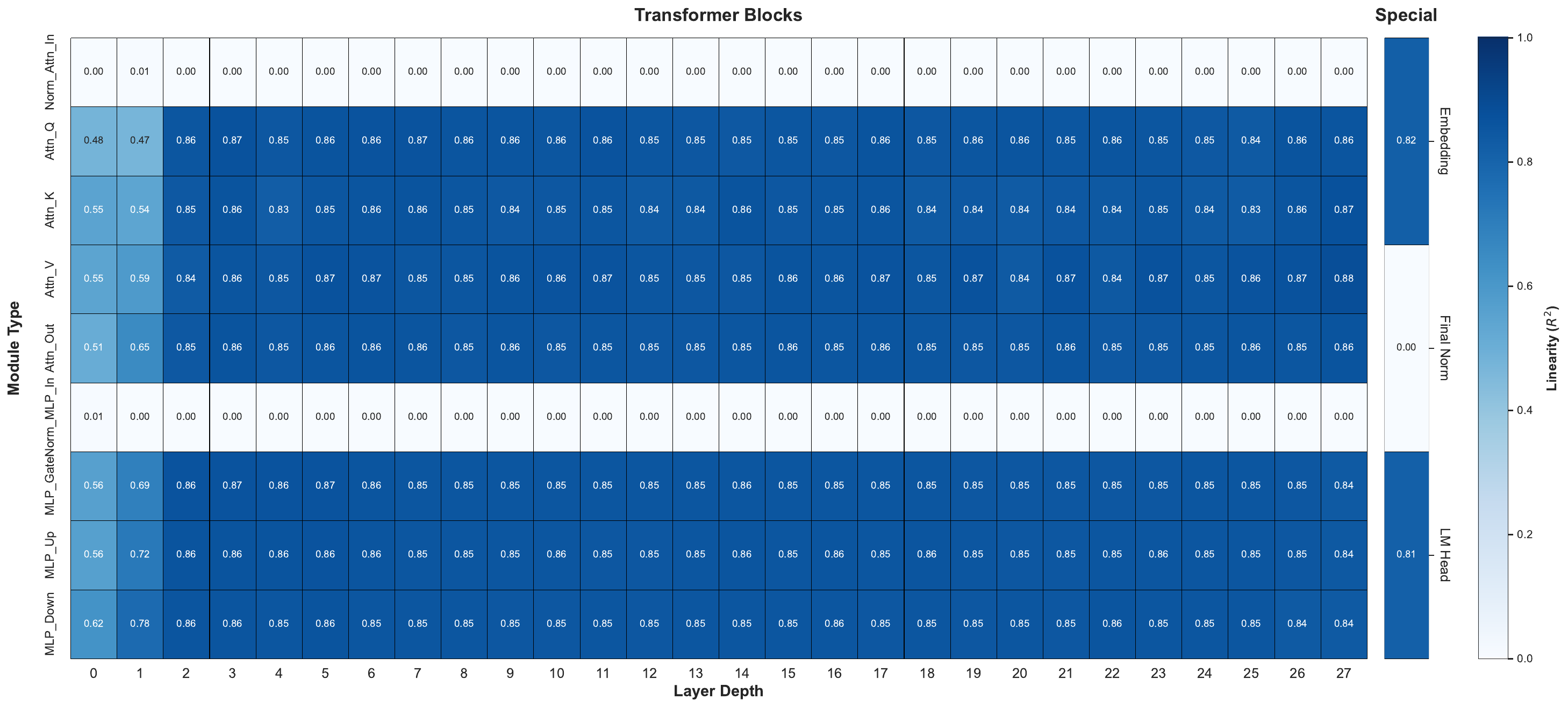}
    \caption[Layer-wise linearity heatmap of model parameters.]{\textbf{Layer-wise linearity heatmap of model parameters.} Each cell reports the mean $R^2$ for a specific module type at a given Transformer layer. Rows correspond to module types, columns correspond to layer indices, and color intensity indicates the magnitude of $R^2$. The separate ``Special'' column summarizes non-block components, including the embedding layer, the final normalization layer, and the language modeling head.}
    \label{fig:weight_r2_layer_heatmap}
\end{figure}

When calculating the $R^2$ to quantify the linearity of weights, we exclude weights that remain constant throughout training, since such trajectories are uninformative about update dynamics.

Figure~\ref{fig:weight_r2_case} illustrates the diverse linear trajectories of weight dynamics during training.
We observe distinct monotonic trends: a subset of weights exhibits a near-linear increase over the course of training, whereas others demonstrate a proportional linear decrease. 

As illustrated in Figure \ref{fig:weight_r2_layer_heatmap}, we analyze the linearity of weight trajectories by calculating the average $R^2$ for each layer.
A notable observation is that all Layer Normalization (LayerNorm) layers exhibit consistently low linearity compared to other layers.

This phenomenon can be attributed to the functional decoupling within the Transformer architecture: LayerNorm parameters primarily govern the statistical normalization of activations, which tends to stabilize early in the training process. In contrast, weights associated with feature transformation (e.g., projection and feed-forward matrices) undergo continuous, linear updates to refine the model's representational capacity.

\subsection{Linearity in Outputs}
\label{app:output_linearity_details}

\paragraph{Details of Output-Space Linearity Measurement}
To evaluate output-space linearity, we select a set of queries from AIME24 and use the base model to generate solution trajectories. Specifically, we sample 64 traces per query using the same generation hyperparameters as in our standard evaluation (Appendix~\ref{app:eval_setup}).
These trajectories are held fixed throughout the analysis to serve as ground-truth sequences.

For each checkpoint, we compute the conditional log-probability of every generated token using teacher-forcing, i.e., given its preceding context within the same fixed probe trajectory.
This yields, for each token position, a trajectory of log-probabilities over training steps.
We then fit a linear regression model to each such trajectory as a function of training steps and compute the corresponding coefficient of determination, $R^2$.

Using fixed probe trajectories under teacher forcing is important for identification: it isolates changes in the model's preference over the same reasoning traces, without conflating them with changes in the sampled outputs themselves.
Thus, the resulting $R^2$ values reflect how linearly the model's token-level preferences evolve during RLVR.

\begin{figure}[t]
    \centering
    \includegraphics[width=\columnwidth]{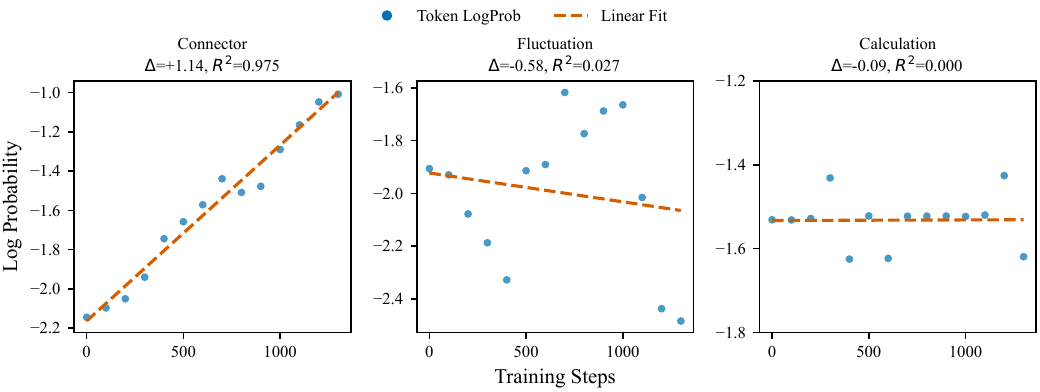}
    \caption[Case study of token log-probability dynamics.]{\textbf{Case study of token log-probability dynamics.} 
    \textbf{Left:} Tokens acting as logical connectors, characterized by significant log-probability changes and high $R^2$ values. 
    \textbf{Middle:} Tokens with large variations in log-probability but low $R^2$, where the probability fluctuates irregularly. 
    \textbf{Right:} Tokens with smaller log-probability variations, which are mostly components of mathematical calculations.}
    \label{fig:token_different_category}
\end{figure}

\paragraph{Token categories by magnitude of change and linearity.}
Beyond the overall $R^2$ distribution, we further analyze how output linearity varies across different types of token trajectories.
We observe a positive association between the magnitude of log-probability change and linearity: tokens with larger behavioral shifts tend to achieve higher $R^2$.
To characterize this pattern more concretely, we group tokens according to their trajectory variance and linearity, as shown in Figure~\ref{fig:token_different_category}.

\begin{itemize}
    \item \textbf{High variance, high $R^2$.} This is the dominant category. It contains many discourse-level reasoning markers, such as ``wait'', ``but'', and ``therefore'', together with nearby continuation tokens. A representative example is the connector phrase ``Wait, that seems a bit messy,'' whose probability increases substantially and does so in a highly linear manner. This indicates that major behavioral shifts during RLVR are often strongly organized.
    \item \textbf{High variance, low $R^2$.} A smaller subset of tokens undergoes substantial change without following a clear linear trend. For example, the continuation ``Similarly, the two circles of radius 1 are tangent to each other and to AB and BC,'' exhibits large fluctuations rather than a monotonic trajectory, reflecting more irregular deviations around the dominant linear pattern.
    \item \textbf{Low variance.} Many tokens change little throughout training and therefore form a stable category. A substantial fraction of these tokens are associated with mathematical calculation content, such as ``$250 \div 188 = 1*188 = 188$, remainder $62$,'' whose log-probabilities remain relatively unchanged during RLVR.
\end{itemize}

Overall, this categorization suggests that output linearity is not uniform across all tokens: it is strongest precisely for those tokens that account for the largest behavioral shifts.

\paragraph{Output-Space Quantities}
\begin{figure}[t]
    \centering
    \includegraphics[width=\columnwidth]{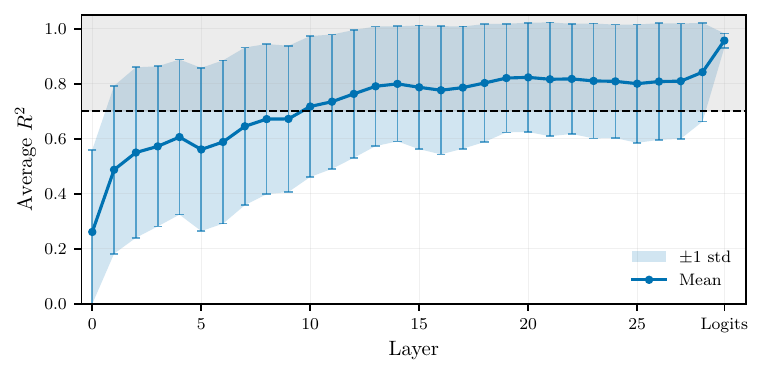}
    \caption[Layer-wise linearity of activations and logits during RL training.]{\textbf{Layer-wise linearity of activations and logits during RL training.} For each layer, we compute the $R^2$ between the RL training step and each neuron's activation value across checkpoints. The solid line shows the mean $R^2$ within each layer, and the shaded region denotes $\pm 1$ standard deviation across selected neurons. The final point corresponds to the logits layer.}
    \label{fig:activation_r2_each_layer}
\end{figure}

To verify if temporal regularity extends beyond token log-probabilities, we analyze the trajectories of logits and intermediate activations across training checkpoints.

Empirical evidence suggests that both quantities exhibit consistent linear trends.
As shown in Figure~\ref{fig:activation_r2_each_layer}, both logits and internal activations exhibit high $R^2$ values ($R^2 \approx 1.0$), indicating a robust linear correlation with training progress.
Specifically, layerwise analysis reveals that all intermediate representations, excluding the initial layer, undergo highly structured drift.
This diminished linearity in the initial layer is likely attributable to its direct coupling with the embedding layer, which experiences complex, high-dimensional fluctuations as the model adapts to input distributions.

The $R^2$ distribution for logits mirrors the concentration observed in log-probabilities, while the linearity of activations suggests that macro-level behavioral regularity is fundamentally rooted in the consistent evolution of internal representations. 
These findings confirm that the linear regime of RLVR is not confined to specific observables but permeates multiple levels of model representation.

\subsection{Robustness of Linearity}\label{app:r2_robustness}

\begin{table}[t]
\centering
\caption{Comprehensive overview of RL training configurations and evaluation results. 
Each row details a distinct experiment with varying models, training datasets, algorithms, 
and hyperparameters. Weight $R^2$ and Token $R^2$ are computed from model weights and 
token log-probabilities, respectively; for each metric, we report the median $R^2$ across 
training checkpoints and $\Pr(R^2 {>} 0.7)$, the proportion of checkpoints exceeding the 
0.7 threshold. The final row reports an SFT baseline for reference; rollouts are not 
applicable (---) in this setting.}
\resizebox{\textwidth}{!}{
\begin{tabular}{lccccccccccc}
\toprule
\multirow{2}{*}{Model}
  & \multirow{2}{*}{Data}
  & \multirow{2}{*}{Algorithm}
  & \multirow{2}{*}{Optimizer}
  & \multirow{2}{*}{LR}
  & \multirow{2}{*}{Batch}
  & \multirow{2}{*}{Rollouts}
  & \multicolumn{2}{c}{Weight $R^2$}
  & \multicolumn{2}{c}{Token $R^2$} \\
\cmidrule(lr){8-9} \cmidrule(lr){10-11}
& & & & & & &
  Median & $\Pr({>}0.7)$ &
  Median & $\Pr({>}0.7)$ \\
\midrule
\multirow{6}{*}{\texttt{DS-R1-Qwen-1.5B}}
  & \multirow{6}{*}{DeepScaleR}
  & GRPO        & AdamW & 1e-6 & 128 & 16 & 0.845 & 0.794 & 0.831 & 0.720 \\
  &
  & GRPO        & AdamW & 3e-6 & 128 & 16 & 0.812 & 0.739 & 0.832 & 0.701 \\
  &
  & GRPO        & AdamW & 1e-5 & 128 & 16 & 0.835 & 0.762 & 0.783 & 0.660 \\
  &
  & REINFORCE++ & AdamW & 1e-6 & 128 & 16 & 0.732 & 0.707 & 0.758 & 0.621 \\
  &
  & GSPO        & AdamW & 1e-6 & 128 & 16 & 0.752 & 0.717 & 0.802 & 0.691 \\
  &
  & GRPO        & SGD   & 1e-6 & 128 & 16 & 0.810 & 0.734 & 0.768 & 0.631 \\
\midrule
\texttt{Nemotron-1.5B}  & DeepScaleR  & GRPO & AdamW & 1e-6 & 256 & 8  & 0.757 & 0.725 & 0.732 & 0.605 \\
\texttt{DS-R1-Qwen-7B}  & Skywork     & GRPO & AdamW & 1e-6 & 256 & 16 & 0.862 & 0.816 & 0.835 & 0.735 \\
\texttt{DS-R1-Llama-8B} & Skywork     & GRPO & AdamW & 1e-6 & 256 & 16 & 0.774 & 0.742 & 0.743 & 0.694 \\
\texttt{Qwen3-8B}       & DAPO-Math   & GRPO & AdamW & 1e-6 & 256 & 8  & 0.868 & 0.824 & 0.829 & 0.711 \\
\texttt{Qwen2.5-32B}    & RLVR-IFeval & GRPO & AdamW & 1e-6 & 128 & 16 & 0.811 & 0.777 & 0.826 & 0.705 \\
\texttt{Qwen3-30B-A3B}  & Skywork     & GSPO & AdamW & 1e-6 & 512 & 16 & 0.750 & 0.702 & 0.781 & 0.637 \\
\midrule
\texttt{Qwen2.5-1.5B}   & GSM8K       & SFT  & AdamW & 1e-3 & 256 & --- & 0.426 & 0.259 & 0.344 & 0.210 \\
\bottomrule
\end{tabular}}
\label{tab:r2_robustness}
\end{table}

Table~\ref{tab:r2_robustness} presents a comprehensive overview of the training configurations and corresponding linearity evaluations. To ensure the optimal RL performance, we utilize intermediate checkpoints from several established projects: DeepScaleR~\citep{deepscaler2025} for \texttt{DeepSeek-R1-Distill-Qwen-1.5B}, JustRL~\citep{he2025justrlscaling15bllm} for \texttt{Open-nemotron-1.5B}, and Skywork-OR1-7B~\citep{he2025skyworkopenreasoner1} for \texttt{DeepSeek-R1-Distill-Qwen-7B}. Furthermore, the instruction-following RLVR dataset is derived from AllenAI's RLVR-IFeval~\citep{lambert2024tulu}, which samples prompts from the Tulu 2 SFT mixture and augments them with verifiable constraints based on the IFEval taxonomy.

\section{Additional Results on Origins of Linearity}
\label{app:origin}
\subsection{Investigation on Optimizers}
Following~\citet{mukherjee2026needadamsurprisinglystrong}, we directly replace AdamW with SGD, setting the learning rate to $[1\text{e-}1, 1\text{e}0]$, momentum to $0$, and weight decay to $0$.
\subsection{Investigation on Zero-RL}
We used the GRPO algorithm to train directly on the base model Qwen3-8B~\citep{yang2025qwen3technicalreport} using DAPO-Math-17k~\citep{NEURIPS2025_a4277440}.

\subsection{Implementation Details of Noise Injection in SFT}
\paragraph{Noise-injected SFT objectives.}
Let
\[
\ell_t = -\log p_\theta(x_{t+1}\mid x_{\le t})
\]
denote the per-token next-token negative log-likelihood, and let \(\tilde m_t \in \{0,1\}\) be the left-shifted SFT mask indicating whether position \(t\) predicts a supervised assistant token. The standard SFT loss is
\[
\mathcal L_{\mathrm{SFT}} = \frac{1}{N}\sum_t \tilde m_t \,\ell_t,
\qquad
N=\sum_t \tilde m_t.
\]

For \textbf{token-level noise injection}, we sample an independent sign for each token,
\[
s_t \in \{+1,-1\}, \qquad P(s_t=-1)=p,
\]
and define
\[
\mathcal L_{\mathrm{token}} = \frac{1}{N}\sum_t \tilde m_t\, s_t\, \ell_t.
\]

For \textbf{sequence-level noise injection}, we sample one cut point \(c\) and one sign \(s\) per sequence,
\[
c \sim \mathrm{Uniform}\{0,1,\dots,T\}, \qquad
s \in \{+1,-1\}, \quad P(s=-1)=p,
\]
and retain only the prefix before \(c\):
\[
q_t = \mathbf{1}[t<c].
\]
The loss is
\[
\mathcal L_{\mathrm{seq}} = \frac{1}{N}\sum_t \tilde m_t\, q_t\, s\, \ell_t.
\]

In both variants, the denominator \(N\) is unchanged from standard SFT, i.e., we always normalize by the original number of valid supervised tokens rather than by the number of retained noisy tokens.

\section{Theoretical Analysis}\label{append:theorem}
We give the formal statement and proof of the theoretical result summarized in Section~\ref{subsec:noisy_signals}.

Let $x \sim \mathcal{D}$ denote the input prompt and $\tau = (y_1, \dots, y_L)$ be the generated response. The policy $\pi_\theta(\tau \mid x)$ defines a probability distribution over the response space $\mathcal{T}$. We consider a binary reward function $R(x, \tau) \in \{0, 1\}$ as a representative case. GRPO incorporates group-relative normalization using the expected reward $\bar{p}_\theta = \mathbb{E}_{x \sim \mathcal{D}, \tau \sim \pi_\theta} [R(x, \tau)]$ and its standard deviation $\sigma_\theta = \sqrt{\bar{p}_\theta(1-\bar{p}_\theta)}$. To focus on the core optimization dynamics, we consider the simplified GRPO objective $J(\theta)$ as the expected normalized reward, omitting the clipping mechanism and KL divergence term:
\begin{equation}
    J(\theta) = \mathbb{E}_{x \sim \mathcal{D}} \left[ \sum_{\tau \in \mathcal{T}} \pi_\theta(\tau \mid x) \left( \frac{R(x, \tau) - \bar{p}_\theta}{\sigma_\theta} \right) \right].
\end{equation}
To analyze the optimization dynamics, we compute the gradient $\nabla_\theta J(\theta)$ using the score function $\varphi_\theta(x, \tau) := \nabla_\theta \log \pi_\theta(\tau \mid x)$. By the log-derivative trick, the gradient is:
\begin{align}
    \nabla_\theta J(\theta) &= \mathbb{E}_{x \sim \mathcal{D}} \left[ \sum_{\tau \in \mathcal{T}} \left( \frac{R(x, \tau) - \bar{p}_\theta}{\sigma_\theta} \right) \nabla_\theta \pi_\theta(\tau \mid x) \right] \nonumber \\
    &= \mathbb{E}_{x \sim \mathcal{D}} \left[ \sum_{\tau \in \mathcal{T}} \pi_\theta(\tau \mid x) \left( \frac{R(x, \tau) - \bar{p}_\theta}{\sigma_\theta} \right) \varphi_\theta(x, \tau) \right] \nonumber \\
    &= \mathbb{E}_{x, \tau \sim \pi_\theta} \left[\left( \frac{R(x, \tau) - \bar{p}_\theta}{\sigma_\theta} \right) \varphi_\theta(x, \tau)\right].
\end{align}

For the remainder of this analysis, we denote $g_{\text{GRPO}}(\theta) = \nabla_\theta J(\theta)$ as the gradient and omit $x$ for brevity. Let $\theta_0$ be the initial parameters and $\delta = \theta - \theta_0$. We adopt the following assumption:

\begin{assumption}[Lazy Training Regime]
There exists a neighborhood $\mathcal{N}_\Omega := \{\theta : \|\theta - \theta_0\| \le \Omega\}$ such that for all $\theta \in \mathcal{N}_\Omega$ and all $\tau \in \mathcal{T}$, the score function satisfies:
\begin{equation}
    \|\varphi_\theta(\tau) - \varphi_{\theta_0}(\tau)\| \le \kappa \|\theta - \theta_0\|,
\end{equation}
where $\kappa \Omega \ll 1$. Let $L = \sup_{\tau \in \mathcal{T}} \|\varphi_{\theta_0}(\tau)\|$ be bounded.
\end{assumption}

This is the standard lazy regime in NTK theory~\citep{jacot2018neural, chizat2019lazy}. Recent empirical work~\citep{malladi2023kernel} indicates that LLM fine-tuning often resides within this regime, justifying the use of a locally stable $\varphi_{\theta_0}$ as a first-order approximation.

\begin{proof}[Proof of Theorem~\ref{main_theorem}]
We first decompose the gradient $g_{\text{GRPO}}(\theta)$ into the score at $\theta_0$ and a drift term:
\begin{equation}
    g_{\text{GRPO}}(\theta) = \mathbb{E}_{\pi_\theta}\left[ \frac{R(\tau) - \bar{p}_\theta}{\sigma_\theta} \varphi_{\theta_0}(\tau) \right] + \mathbb{E}_{\pi_\theta}\left[ \frac{R(\tau) - \bar{p}_\theta}{\sigma_\theta} (\varphi_\theta(\tau) - \varphi_{\theta_0}(\tau)) \right].
\end{equation}
Since $|R(\tau) - \bar{p}_\theta| \le 1$, the second term is bounded by $\frac{\kappa}{\sigma_\theta} \|\delta\|$. Defining the success set $S = \{\tau : R(\tau) = 1\}$, the first term becomes:
\begin{align}
    \mathbb{E}_{\pi_\theta}\left[ \frac{R(\tau) - \bar{p}_\theta}{\sigma_\theta} \varphi_{\theta_0} \right] &= \frac{1}{\sigma_\theta} \left( \sum_{\tau \in S} \pi_\theta(\tau) \varphi_{\theta_0}(\tau) - \bar{p}_\theta \sum_{\tau \in \mathcal{T}} \pi_\theta(\tau) \varphi_{\theta_0}(\tau) \right) \nonumber \\
    &= \frac{1}{\sigma_\theta} \left( \sum_{\tau \in S} \pi_\theta(\tau) \varphi_{\theta_0}(\tau) - \bar{p}_\theta \sum_{\tau \in \mathcal{T}} \pi_\theta(\tau) (\varphi_{\theta_0}(\tau) - \varphi_\theta(\tau)) \right),
\end{align}
where we used $\sum_{\tau \in \mathcal{T}} \pi_\theta(\tau) \varphi_\theta(\tau) = 0$. The sum over $\mathcal{T}$ is bounded by $\kappa \|\delta\|$, leading to $g_{\text{GRPO}}(\theta) = \frac{1}{\sigma_\theta} \sum_{\tau \in S} \pi_\theta(\tau) \varphi_{\theta_0}(\tau) + r_1$, with $\|r_1\| \le \frac{2\kappa}{\sigma_\theta} \|\delta\|$.

To evaluate the sum over $S$, we expand the density ratio. For any response $\tau$, we consider the straight-line path $\theta_s := \theta_0 + s\delta$ for $s \in [0, 1]$. Then, the log-density ratio satisfies:
\begin{equation}\label{eq:expand_xr}
    \log \pi_\theta(\tau) - \log \pi_{\theta_0}(\tau) = \int_0^1 \langle \delta, \varphi_{\theta_s}(\tau) \rangle ds.
\end{equation}
Applying Assumption 1, the integrand satisfies $\langle \delta, \varphi_{\theta_s}(\tau) \rangle = \langle \delta, \varphi_{\theta_0}(\tau) \rangle + \epsilon_s$ where $|\epsilon_s| \le \kappa s \|\delta\|^2$. Integrating over $s \in [0, 1]$, we obtain $\log (\pi_\theta / \pi_{\theta_0}) = \langle \delta, \varphi_{\theta_0}(\tau) \rangle + \epsilon_\tau$ with $|\epsilon_\tau| \le \frac{\kappa}{2}\|\delta\|^2$. Consequently, $\pi_\theta(\tau) = \pi_{\theta_0}(\tau) (1 + \langle \delta, \varphi_{\theta_0}(\tau) \rangle + \rho_\tau)$ where $|\rho_\tau| \le C \|\delta\|^2$ follows from the Taylor expansion of the exponential function. $C$ is determined by the maximum curvature of the log-likelihood in $\mathcal{N}_\Omega$ (detailed in Appendix~\ref{derivation_C}). Substituting this into the sum over $S$:
\begin{align}
    \sum_{\tau \in S} \pi_\theta(\tau) \varphi_{\theta_0}(\tau) &= \sum_{\tau \in S} \pi_{\theta_0}(\tau) \varphi_{\theta_0}(\tau) + \left( \sum_{\tau \in S} \pi_{\theta_0}(\tau) \varphi_{\theta_0}(\tau) \varphi_{\theta_0}(\tau)^\top \right) \delta + \sum_{\tau \in S} \pi_{\theta_0}(\tau) \rho_\tau \varphi_{\theta_0}(\tau) \nonumber \\
    &= \bar{p}_0 v_R + \bar{p}_0 M_S \delta + r_2,
\end{align}
where $\bar{p}_0 = \Pr_{\theta_0}(S)$, $M_S = \mathbb{E}_{\pi_{\theta_0}}[\varphi_{\theta_0} \varphi_{\theta_0}^\top \mid S]$, and $\|r_2\| \le C L \bar{p}_0 \|\delta\|^2$. Similarly, expanding $\bar{p}_\theta$ yields:
\begin{equation}
\bar{p}_\theta = \sum_{\tau \in S} \pi_\theta(\tau) = \bar{p}_0 + \bar{p}_0 v_R^\top \delta + r_3, \quad \text{where } |r_3| \le C \bar{p}_0 \|\delta\|^2.
\end{equation}
Combining these to analyze the deviation from the direction $\frac{\bar{p}_\theta}{\sigma_\theta} v_R$:
\begin{align}
    g_{\text{GRPO}}(\theta) - \frac{\bar{p}_\theta}{\sigma_\theta} v_R &= \frac{1}{\sigma_\theta} (\bar{p}_0 v_R + \bar{p}_0 M_S \delta + r_2) - \frac{1}{\sigma_\theta} (\bar{p}_0 + \bar{p}_0 v_R^\top \delta + r_3) v_R + r_1 \nonumber \\
    &= \frac{\bar{p}_0}{\sigma_\theta} (M_S - v_R v_R^\top) \delta + \frac{r_2}{\sigma_\theta} - \frac{r_3 v_R}{\sigma_\theta} + r_1.
\end{align}
Defining $K_S = M_S - v_R v_R^\top$, the norm of the deviation is:
\begin{equation}
    \| g_{\text{GRPO}}(\theta) - \frac{\bar{p}_\theta}{\sigma_\theta} v_R \| \le \frac{\bar{p}_0}{\sigma_\theta} \|K_S\|_{\text{op}} \|\delta\| + \frac{2\kappa}{\sigma_\theta} \|\delta\| + \frac{1}{\sigma_\theta} (C L + C \|v_R\|) \bar{p}_0 \|\delta\|^2.
\end{equation}
Since $\bar{p}_0 \le 1$ and $\|v_R\| \le L$, the conclusion holds.
\end{proof}

\subsection{Detailed Derivation of $C$}\label{derivation_C}
In this section, we provide a formal derivation of the second-order constant $C$ used in Theorem 1. Our objective is to bound the residual $\rho_\tau$ in the density ratio expansion $\pi_\theta(\tau)/\pi_{\theta_0}(\tau) = 1 + \varphi_{\theta_0}(\tau)^\top \delta + \rho_\tau$.

\paragraph{D.1.1 Taylor Expansion with Lagrange Remainder.} 
Consider the log-density difference $x_\tau := \log \pi_\theta(\tau) - \log \pi_{\theta_0}(\tau)$. By the second-order Taylor theorem, there exists a point $\xi_\tau$ strictly between $0$ and $x_\tau$ such that:
\begin{equation}
    e^{x_\tau} = 1 + x_\tau + \frac{1}{2}x_\tau^2 e^{\xi_\tau}.
\end{equation}
The term $e^{\xi_\tau}$ captures the local curvature of the exponential mapping. In the lazy training regime, we operate within the neighborhood $\mathcal{N}_\Omega$, where the displacement $\|\delta\| \le \Omega$ is small. Since $|x_\tau| \le L\|\delta\| + \frac{\kappa}{2}\|\delta\|^2$, it follows that $x_\tau \to 0$ as $\Omega \to 0$, implying $e^{\xi_\tau}$ is well-behaved and close to unity. We define a uniform upper bound for this curvature across $\mathcal{N}_\Omega$ as:
\begin{equation}
    e^\eta := \sup_{\theta \in \mathcal{N}_\Omega, \tau \in \mathcal{T}} e^{\xi_\tau} \le \exp\left( L\Omega + \frac{\kappa}{2}\Omega^2 \right).
\end{equation}

\paragraph{D.1.2 Bounding the Pointwise Residual $\rho_\tau$.}
Combining the expansion with our control on $x_\tau$ in Eq.~\eqref{eq:expand_xr}, where $x_\tau = \varphi_{\theta_0}(\tau)^\top \delta + \epsilon_\tau$ with $|\epsilon_\tau| \le \frac{\kappa}{2}\|\delta\|^2$, we have:
\begin{equation}
    \rho_\tau = \epsilon_\tau + \frac{1}{2}\left( \varphi_{\theta_0}(\tau)^\top \delta + \epsilon_\tau \right)^2 e^{\xi_\tau}.
\end{equation}
Taking absolute values and applying the triangle inequality:
\begin{align}
    |\rho_\tau| &\le |\epsilon_\tau| + \frac{1}{2}\left( |\varphi_{\theta_0}(\tau)^\top \delta| + |\epsilon_\tau| \right)^2 e^{\xi_\tau} \nonumber \\
    &\le \frac{\kappa}{2}\|\delta\|^2 + \frac{1}{2}\left( L\|\delta\| + \frac{\kappa}{2}\|\delta\|^2 \right)^2 e^\eta.
\end{align}
Neglecting the higher-order terms $\mathcal{O}(\|\delta\|^3)$, the residual is bounded by:
\begin{equation}
    |\rho_\tau| \le \frac{1}{2} (L^2 e^\eta + \kappa) \|\delta\|^2.
\end{equation}

\paragraph{D.1.3 Explicit Definition of $C$.}
Comparing with the bound $|\rho_\tau| \le C \|\delta\|^2$ used in the proof of Theorem 1, the constant $C$ is explicitly given by:
\begin{equation}
    C := \frac{1}{2} \left( L^2 e^\eta + \kappa \right),
\end{equation}
where $e^\eta \le \exp\left( L\Omega + \frac{\kappa}{2}\Omega^2 \right)$ uniformly bounds the exponential curvature in $\mathcal{N}_\Omega$. In the lazy training regime where $\kappa\Omega \ll 1$ and $L\Omega$ is moderate, $e^\eta = 1 + \mathcal{O}(L\Omega)$, so $C$ is a finite constant depending only on the three external quantities $L$, $\Omega$, and $\kappa$, and is independent of the model dimension $d$. In particular, when $\Omega \to 0$ we recover $C \to \frac{1}{2}(L^2 + \kappa)$.

\section{Additional Results on Extrapolation}
\subsection{Empirical Guidelines for Weight-space Extrapolation Hyperparameters}
\label{app:weight_extra}
Based on our extensive experimental findings, the performance of weight-space extrapolation relies on the appropriate selection of the anchor checkpoints ($t_0$ and $t_1$) and the extrapolation ratio ($\beta$). Assuming a total available training budget of $N$ steps, we summarize our empirical guidelines for these hyperparameters as follows:

\begin{itemize}[leftmargin=*]
    \item \textbf{Initial Anchor $t_0$:} The model performance is generally not highly sensitive to the exact choice of the starting step $t_0$. In practice, we recommend setting $t_0$ to approximately $20\%$ to $30\%$ of the total budget $N$. This allows the model to bypass the initial unstable phase of training and provides a reliable starting point for estimating the optimization trajectory.
    
    \item \textbf{Second Anchor $t_1$:} To capture the most mature weight update direction and maximize the utilization of the available training dynamics, $t_1$ should be set as late as possible. Therefore, we recommend setting $t_1 = N$, utilizing the final checkpoint of the training budget.
    
    \item \textbf{Extrapolation Ratio $\beta$:} The optimal scaling factor $\beta$ (which directly determines the target step $t$) is highly task-specific. As discussed in Section~\ref{sec:weight_extra} and illustrated in Figure~\ref{fig:extrapolation}, while a moderate $\beta$ strictly improves performance, an excessively large $\beta$ can amplify trajectory estimation errors. Consequently, $\beta$ should be carefully tuned on a validation set according to the specific downstream task and dataset characteristics.
\end{itemize}

\subsection{Periodic Re-grounding}\label{app:regrounding}
Periodic Re-grounding mechanism serves as an implementation strategy for refreshing the local direction of the trajectory once extrapolation becomes unreliable. Concretely, we alternate between short segments of standard RL optimization and extrapolation-based advancement. 
This design preserves the efficiency benefits of extrapolation while reducing the error accumulation associated with long-horizon projection. The specific configurations (parameters $m$ and $n$) selected for each training budget in Section~\ref{sec:periodic-re-grounding} are listed in Table~\ref{tab:regrounding_hyperparams}.

\paragraph{Difference between Periodic Re-grounding and Fine-tuning LR:} We investigate the fundamental differences between our proposed Periodic Re-grounding mechanism and the naive approach of scaling the learning rate (LR).
Naive LR scaling inevitably fails due to severe gradient noise; specifically, the gradients exhibit a critically low correlation (< 0.03) with the dominant update direction and suffer from high variance. Consequently, increasing the learning rate directly amplifies this noise, leading to extreme training instability and frequent crashes.
Empirically, scaling the LR from 1e-6 to 1e-5 results in a severe performance degradation, with AIME24 accuracy drastically dropping from 0.42 to 0.32.
To overcome these limitations, Periodic Re-grounding mechanism achieves stable acceleration through an identify-then-extrapolate strategy. Rather than amplifying noisy updates, it first calibrates the model using 20 actual reinforcement learning (RL) steps to reliably establish a stable linear direction. Following this calibration, it executes a long-step extrapolation by projecting 100 steps along this identified trajectory.
This approach effectively circumvents gradient noise, yielding a >6x training speedup while strictly preserving model accuracy.

\begin{table}[t]
    \centering
\caption{\textbf{Periodic Re-grounding Hyperparameter Configurations.} 
Values are reported for each training budget $s$ corresponding to Table~\ref{tab:regrounding_performance}.}
\label{tab:regrounding_hyperparams}
\setlength{\tabcolsep}{12pt}
\begin{tabular}{c c c}
\toprule
\textbf{Training Budget ($s$)} & \textbf{$m$} & \textbf{$n$} \\
\midrule
200   & 100 & 100 \\
400   & 300 & 600 \\
800   & 100 & 100 \\
1200  & 100 & 100 \\
\bottomrule
\end{tabular}
\end{table}

\section{Implications for Understanding RLVR}\label{sec:implication}
\paragraph{Elicitation vs. Emergence in RLVR.}
A central debate in reasoning-oriented LLMs concerns whether RLVR instills novel cognitive capabilities or merely elicits pre-existing ones.
Proponents of capability emergence argue that verifiable rewards facilitate the composition of atomic reasoning primitives to solve unseen tasks~\citep{wen2025reinforcementlearningverifiablerewards, wang2026newskillssharperprimitives}.
Conversely, the ``Invisible Leash'' hypothesis~\citep{wu2026invisibleleashrlvrescape,NEURIPS2025_537d5aa7} contends that RLVR is strictly bounded by the base model's pre-trained manifold. This elicitation-only view is empirically supported by test-time scaling studies, where majority voting often matches RLVR peak performance~\citep{wang2022self, karan2026reasoning}.
Our mechanistic findings on RLVR linearity lend rigorous mathematical support to the elicitation hypothesis: rather than executing complex nonlinear searches for novel behaviors, current RLVR algorithms predominantly amplify and up-weight dominant reasoning trajectories established in the earliest stages of training.

This perspective also suggests a concrete direction for future RLVR algorithms.
RL training must promote reasoning patterns that are initially rare under the base model. New skills are more likely to arise when low-probability but valuable traces are surfaced and selectively reinforced.
This could be achieved by expanding exploration during rollout, e.g., with higher temperatures and larger rollout budgets, together with reweighting schemes that favor novel successful samples, or by replaying rare positive traces that standard on-policy updates might otherwise wash out.
More broadly, escaping the current linear, elicitation-dominated regime may require RLVR algorithms that explicitly combine exploration, selective amplification, and memory.

\paragraph{Training Stability \& Acceleration.}
Weight linearity also suggests a promising route toward more efficient RLVR optimization.
If training follows a linear trajectory in parameter space, then small updates obtained from different batches and successive steps may be safely accumulated into a much larger effective gradient update. This opens up the possibility of replacing many fine-grained parameter updates with less frequent but larger, more compute-efficient ones.
From this perspective, a key research question is to characterize the scaling law among learning rate, batch size, and rollout budget: understanding how these quantities co-vary within the linear regime may reveal operating points that simultaneously improve stability and accelerate training.
More broadly, the linear structure we observe suggests that RLVR may be amenable to principled update accumulation strategies that better exploit available compute without sacrificing optimization reliability.

\paragraph{Instability in the Later Stages.}
As shown in Table~\ref{tab:extrapolation}, output-space extrapolation outperforms standard RL across three benchmarks.
A plausible explanation is that output-space extrapolation preserves the stable improvement direction established in earlier training, while avoiding some of the degradation that can arise in late-stage RLVR, such as entropy collapse or over-fitting.
In this sense, output-space extrapolation acts as a lightweight way to continue the policy's functional trajectory without incurring the full cost---or instability---of additional RL updates.

Our results also suggest a new perspective on RLVR instability.
In our experiments, weight evolution loses linearity near entropy collapse.
This raises two possibilities: either continued logit amplification drives entropy collapse and then breaks the linear regime, or linearity breaks first and serves as an early warning signal for instability. In the latter case, extrapolation could be not only predictive but also stabilizing, by keeping training on the pre-collapse trajectory and away from unstable regions.
This points to a broader direction: understanding the interplay between logit growth, entropy dynamics, and trajectory linearity may help anticipate and mitigate RLVR instability.

\section{Limitations}\label{sec:limitation}
While our study provides robust empirical and theoretical evidence for the linear regime in RLVR, it presents several limitations that point to promising directions for future research:

\begin{itemize}[leftmargin=*]
    \item \textbf{Scope of Reward Modalities:} Our analysis strictly focuses on Reinforcement Learning with Verifiable Rewards (RLVR) in domains with objective correctness, such as mathematics and coding. It remains an open question whether similar striking linear dynamics dominate in standard RLHF pipelines that rely on learned, noisy reward models, or in open-ended text generation tasks where the reward landscape is fundamentally different.
    
    \item \textbf{Theoretical Assumptions:} Our mechanistic explanation and formal analysis (Theorem~\ref{main_theorem}) rely on an NTK-style lazy training assumption. While this is empirically justified for the fine-tuning phases and learning rates observed in current RLVR practices, this assumption may eventually break down during extremely long-horizon optimization, or if the base model undergoes a severe distribution shift that forces it out of the locally linear neighborhood.
    
    \item \textbf{Extrapolation Horizon and Hyperparameter Tuning:} Although weight-space extrapolation yields significant compute savings, the optimal extrapolation ratio ($\beta$) remains highly task-specific. As demonstrated, pushing the projection horizon too far without periodic re-grounding amplifies trajectory estimation errors. Consequently, identifying the optimal hyperparameter configuration currently requires empirical tuning on a validation set, adding slight overhead to the otherwise compute-saving method.
    
    \item \textbf{Model Scale and Algorithms:} We evaluated models up to 32B parameters and focused primarily on memory-efficient, critic-free RL algorithms (e.g., GRPO, REINFORCE++, and GSPO). While the linear regime persists consistently across these configurations, verifying these dynamics on frontier-scale models (e.g., 100B+ parameters) or under standard actor-critic PPO frameworks is deferred to future work due to computational constraints.

    \item \textbf{Curriculum Learning:} Our analysis (Theorem \ref{main_theorem}) assumes a fixed prompt distribution $\mathcal{D}$. Under Curriculum Learning, the dynamic evolution of the data distribution $\mathcal{D}_t$ may shift the dominant optimization subspace. We hypothesize that such trajectories exhibit \textit{piecewise linearity} within stable curriculum stages rather than strict global linearity. Characterizing this geometry under non-stationary distributions remains an important direction for future work.
\end{itemize}

%% file: refs.bib
@article{Guo_2025,
   title={DeepSeek-R1 incentivizes reasoning in LLMs through reinforcement learning},
   journal={Nature},
   volume={645},
   number={8081},
   pages={633--638},
   year={2025},
   month=sept,
   doi={10.1038/s41586-025-09422-z},
   url={http://dx.doi.org/10.1038/s41586-025-09422-z},
   author={Guo, Daya and Yang, Dejian and Zhang, Haowei and Song, Junxiao and Wang, Peiyi and others}
}

@misc{shao2024deepseekmathpushinglimitsmathematical,
      title={DeepSeekMath: Pushing the Limits of Mathematical Reasoning in Open Language Models}, 
      author={Zhihong Shao and Peiyi Wang and Qihao Zhu and Runxin Xu and Junxiao Song and Xiao Bi and Haowei Zhang and Mingchuan Zhang and Y. K. Li and Y. Wu and Daya Guo},
      year={2024},
      eprint={2402.03300},
      archivePrefix={arXiv},
      primaryClass={cs.CL},
      url={https://arxiv.org/abs/2402.03300}, 
}

@misc{deepscaler2025,
  title={DeepScaleR: Surpassing O1-Preview with a 1.5B Model by Scaling RL},
  author={Michael Luo and Sijun Tan and Justin Wong and Xiaoxiang Shi and William Y. Tang and Manan Roongta and Colin Cai and Jeffrey Luo and Li Erran Li and Raluca Ada Popa and Ion Stoica},
  year={2025},
  howpublished={Notion Blog},
  note={Available at \texttt{pretty-radio-b75.notion.site}},
  note={Notion Blog},
  year={2025}
}

@inproceedings{NEURIPS2025_bf235a1d,
 author = {Mukherjee, Sagnik and Yuan, Lifan and Hakkani-Tur, Dilek and Peng, Hao},
 booktitle = {Advances in Neural Information Processing Systems},
 editor = {D. Belgrave and C. Zhang and H. Lin and R. Pascanu and P. Koniusz and M. Ghassemi and N. Chen},
 pages = {132119--132138},
 publisher = {Curran Associates, Inc.},
 title = {Reinforcement Learning Finetunes Small Subnetworks in Large Language Models},
 url = {https://proceedings.neurips.cc/paper_files/paper/2025/file/bf235a1d6780afd979f2f81676f43413-Paper-Conference.pdf},
 volume = {38},
 year = {2025}
}

@inproceedings{NEURIPS2025_537d5aa7,
 author = {Chen, Zhiqi and Lu, Rui and Zhao, Andrew and Wang, Zhaokai and Yue, Yang and Song, Shiji and Huang, Gao},
 booktitle = {Advances in Neural Information Processing Systems},
 editor = {D. Belgrave and C. Zhang and H. Lin and R. Pascanu and P. Koniusz and M. Ghassemi and N. Chen},
 pages = {57654--57689},
 publisher = {Curran Associates, Inc.},
 title = {Does Reinforcement Learning Really Incentivize Reasoning Capacity in LLMs Beyond the Base Model?},
 url = {https://proceedings.neurips.cc/paper_files/paper/2025/file/537d5aa768c2d534016a4d06f87bc8fb-Paper-Conference.pdf},
 volume = {38},
 year = {2025}
}

@misc{wu2026invisibleleashrlvrescape,
      title={The Invisible Leash: Why RLVR May or May Not Escape Its Origin}, 
      author={Fang Wu and Weihao Xuan and Ximing Lu and Mingjie Liu and Yi Dong and Zaid Harchaoui and Yejin Choi},
      year={2026},
      eprint={2507.14843},
      archivePrefix={arXiv},
      primaryClass={cs.LG},
      url={https://arxiv.org/abs/2507.14843}, 
}

@inproceedings{
karan2026reasoning,
title={Reasoning with Sampling: Your Base Model is Smarter Than You Think},
author={Aayush Karan and Yilun Du},
booktitle={The Fourteenth International Conference on Learning Representations},
year={2026},
url={https://openreview.net/forum?id=Vsgq2ldr4K}
}

@inproceedings{
zhu2025the,
title={The Path Not Taken: {RLVR} Provably Learns Off the Principals},
author={Hanqing Zhu and Zhenyu Zhang and Hanxian Huang and DiJia Su and Zechun Liu and Jiawei Zhao and Igor Fedorov and Hamed Pirsiavash and Jinwon Lee and David Z. Pan and Zhangyang Wang and Yuandong Tian and Kai Sheng Tai},
booktitle={NeurIPS 2025 Workshop on Efficient Reasoning},
year={2025},
url={https://openreview.net/forum?id=N75EWQQnb3}
}

@inproceedings{
wang2026reinforcement,
title={Reinforcement Learning for Reasoning in Large Language Models with One Training Example},
author={Yiping Wang and Qing Yang and Zhiyuan Zeng and Liliang Ren and Liyuan Liu and Baolin Peng and Hao Cheng and Xuehai He and Kuan Wang and Jianfeng Gao and Weizhu Chen and Shuohang Wang and Simon Shaolei Du and yelong shen},
booktitle={The Thirty-ninth Annual Conference on Neural Information Processing Systems},
year={2026},
url={https://openreview.net/forum?id=IBrRNLr6JA}
}

@misc{hu2025reinforcestabilizingcriticfreepolicy,
      title={REINFORCE++: Stabilizing Critic-Free Policy Optimization with Global Advantage Normalization}, 
      author={Jian Hu and Jason Klein Liu and Haotian Xu and Wei Shen},
      year={2025},
      eprint={2501.03262},
      archivePrefix={arXiv},
      primaryClass={cs.CL},
      url={https://arxiv.org/abs/2501.03262}, 
}

@inproceedings{NEURIPS2025_a4277440,
 author = {Yu, Qiying and Zhang, Zheng and Zhu, Ruofei and others},
 booktitle = {Advances in Neural Information Processing Systems},
 editor = {D. Belgrave and C. Zhang and H. Lin and R. Pascanu and P. Koniusz and M. Ghassemi and N. Chen},
 pages = {113222--113244},
 publisher = {Curran Associates, Inc.},
 title = {DAPO: An Open-Source LLM Reinforcement Learning System at Scale},
 url = {https://proceedings.neurips.cc/paper_files/paper/2025/file/a4277440d50f1f15d2cb4c14f7e0c0d2-Paper-Conference.pdf},
 volume = {38},
 year = {2025}
}

@misc{zheng2025groupsequencepolicyoptimization,
      title={Group Sequence Policy Optimization}, 
      author={Chujie Zheng and Shixuan Liu and Mingze Li and Xiong-Hui Chen and Bowen Yu and Chang Gao and Kai Dang and Yuqiong Liu and Rui Men and An Yang and Jingren Zhou and Junyang Lin},
      year={2025},
      eprint={2507.18071},
      archivePrefix={arXiv},
      primaryClass={cs.LG},
      url={https://arxiv.org/abs/2507.18071}, 
}

@misc{yang2025qwen3technicalreport,
      title={Qwen3 Technical Report}, 
      author={An Yang and Anfeng Li and Baosong Yang and others},
      year={2025},
      eprint={2505.09388},
      archivePrefix={arXiv},
      primaryClass={cs.CL},
      url={https://arxiv.org/abs/2505.09388}, 
}

@misc{he2025skyworkopenreasoner1,
      title={Skywork Open Reasoner 1 Technical Report}, 
      author={Jujie He and Jiacai Liu and Chris Yuhao Liu and Rui Yan and Chaojie Wang and Peng Cheng and Xiaoyu Zhang and Fuxiang Zhang and Jiacheng Xu and Wei Shen and Siyuan Li and Liang Zeng and Tianwen Wei and Cheng Cheng and Bo An and Yang Liu and Yahui Zhou},
      year={2025},
      eprint={2505.22312},
      archivePrefix={arXiv},
      primaryClass={cs.LG},
      url={https://arxiv.org/abs/2505.22312}, 
}

@misc{moshkov2025aimo2winningsolutionbuilding,
      title={AIMO-2 Winning Solution: Building State-of-the-Art Mathematical Reasoning Models with OpenMathReasoning dataset}, 
      author={Ivan Moshkov and Darragh Hanley and Ivan Sorokin and Shubham Toshniwal and Christof Henkel and Benedikt Schifferer and Wei Du and Igor Gitman},
      year={2025},
      eprint={2504.16891},
      archivePrefix={arXiv},
      primaryClass={cs.AI},
      url={https://arxiv.org/abs/2504.16891}, 
}

@inproceedings{NEURIPS2022_8636419d,
 author = {Le, Hung and Wang, Yue and Gotmare, Akhilesh Deepak and Savarese, Silvio and Hoi, Steven Chu Hong},
 booktitle = {Advances in Neural Information Processing Systems},
 editor = {S. Koyejo and S. Mohamed and A. Agarwal and D. Belgrave and K. Cho and A. Oh},
 pages = {21314--21328},
 publisher = {Curran Associates, Inc.},
 title = {CodeRL: Mastering Code Generation through Pretrained Models and Deep Reinforcement Learning},
 url = {https://proceedings.neurips.cc/paper_files/paper/2022/file/8636419dea1aa9fbd25fc4248e702da4-Paper-Conference.pdf},
 volume = {35},
 year = {2022}
}

@misc{aime24,
  author    = {Yifan Zhang and {Team Math-AI}},
  title     = {American Invitational Mathematics Examination (AIME) 2024}, 
  year      = {2024}
}

@misc{aime25,
  author    = {Yifan Zhang and {Team Math-AI}},
  title     = {American Invitational Mathematics Examination (AIME) 2025}, 
  year      = {2025}
}

@inproceedings{math500,
 author = {Lightman, Hunter and Kosaraju, Vineet and Burda, Yuri and Edwards, Harrison and Baker, Bowen and Lee, Teddy and Leike, Jan and Schulman, John  and Sutskever, Ilya and Cobbe, Karl},
 booktitle = {International Conference on Learning Representations},
 editor = {B. Kim and Y. Yue and S. Chaudhuri and K. Fragkiadaki and M. Khan and Y. Sun},
 pages = {39578--39601},
 title = {Let\textquotesingle s Verify Step by Step},
 url = {https://proceedings.iclr.cc/paper_files/paper/2024/file/aca97732e30bcf1303bc22ac3924fd16-Paper-Conference.pdf},
 volume = {2024},
 year = {2024}
}

@inproceedings{
livecodebench,
title={LiveCodeBench: Holistic and Contamination Free Evaluation of Large Language Models for Code},
author={Naman Jain and King Han and Alex Gu and Wen-Ding Li and Fanjia Yan and Tianjun Zhang and Sida Wang and Armando Solar-Lezama and Koushik Sen and Ion Stoica},
booktitle={The Thirteenth International Conference on Learning Representations},
year={2025},
url={https://openreview.net/forum?id=chfJJYC3iL}
}

@misc{jaech2024openai,
      title={OpenAI o1 System Card}, 
      author={OpenAI and Aaron Jaech and Adam Kalai and others},
      year={2026},
      eprint={2412.16720},
      archivePrefix={arXiv},
      primaryClass={cs.AI},
      url={https://arxiv.org/abs/2412.16720}
}

@inproceedings{lambert2024tulu,
      title={Tulu 3: Pushing Frontiers in Open Language Model Post-Training},
      author={Nathan Lambert and Jacob Morrison and Valentina Pyatkin and others},
      booktitle={Second Conference on Language Modeling},
      year={2025},
      url={https://openreview.net/forum?id=i1uGbfHHpH}
}

@inproceedings{
wang2022self,
title={Self-Consistency Improves Chain of Thought Reasoning in Language Models},
author={Xuezhi Wang and Jason Wei and Dale Schuurmans and Quoc V Le and Ed H. Chi and Sharan Narang and Aakanksha Chowdhery and Denny Zhou},
booktitle={The Eleventh International Conference on Learning Representations },
year={2023},
url={https://openreview.net/forum?id=1PL1NIMMrw}
}

@misc{schulman2017proximal,
      title={Proximal Policy Optimization Algorithms}, 
      author={John Schulman and Filip Wolski and Prafulla Dhariwal and Alec Radford and Oleg Klimov},
      year={2017},
      eprint={1707.06347},
      archivePrefix={arXiv},
      primaryClass={cs.LG},
      url={https://arxiv.org/abs/1707.06347}, 
}

@inproceedings{
wen2025reinforcementlearningverifiablerewards,
title={Reinforcement Learning with Verifiable Rewards Implicitly Incentivizes Correct Reasoning in Base {LLM}s},
author={Xumeng Wen and Zihan Liu and Shun Zheng and Shengyu Ye and Zhirong Wu and Yang Wang and Zhijian Xu and Xiao Liang and Junjie Li and Ziming Miao and Jiang Bian and Mao Yang},
booktitle={The Fourteenth International Conference on Learning Representations},
year={2026},
url={https://openreview.net/forum?id=jGbRWwIidy}
}

@misc{wang2026newskillssharperprimitives,
      title={New Skills or Sharper Primitives? A Probabilistic Perspective on the Emergence of Reasoning in RLVR}, 
      author={Zhilin Wang and Yafu Li and Shunkai Zhang and Zhi Wang and Haoran Zhang and Xiaoye Qu and Yu Cheng},
      year={2026},
      eprint={2602.08281},
      archivePrefix={arXiv},
      primaryClass={cs.CL},
      url={https://arxiv.org/abs/2602.08281}, 
}

@InProceedings{wortsman2022modelsoupsaveragingweights,
  title = 	 {Model soups: averaging weights of multiple fine-tuned models improves accuracy without increasing inference time},
  author =       {Wortsman, Mitchell and Ilharco, Gabriel and Gadre, Samir Ya and Roelofs, Rebecca and Gontijo-Lopes, Raphael and Morcos, Ari S and Namkoong, Hongseok and Farhadi, Ali and Carmon, Yair and Kornblith, Simon and Schmidt, Ludwig},
  booktitle = 	 {Proceedings of the 39th International Conference on Machine Learning},
  pages = 	 {23965--23998},
  year = 	 {2022},
  editor = 	 {Chaudhuri, Kamalika and Jegelka, Stefanie and Song, Le and Szepesvari, Csaba and Niu, Gang and Sabato, Sivan},
  volume = 	 {162},
  series = 	 {Proceedings of Machine Learning Research},
  month = 	 {17--23 Jul},
  publisher =    {PMLR},
  url = 	 {https://proceedings.mlr.press/v162/wortsman22a.html}
}

@inproceedings{zheng2025modelextrapolationexpeditesalignment,
    title = "Model Extrapolation Expedites Alignment",
    author = "Zheng, Chujie  and
      Wang, Ziqi  and
      Ji, Heng  and
      Huang, Minlie  and
      Peng, Nanyun",
    editor = "Che, Wanxiang  and
      Nabende, Joyce  and
      Shutova, Ekaterina  and
      Pilehvar, Mohammad Taher",
    booktitle = "Proceedings of the 63rd Annual Meeting of the Association for Computational Linguistics (Volume 1: Long Papers)",
    month = jul,
    year = "2025",
    address = "Vienna, Austria",
    publisher = "Association for Computational Linguistics",
    url = "https://aclanthology.org/2025.acl-long.51/",
    doi = "10.18653/v1/2025.acl-long.51",
    pages = "1025--1041",
    ISBN = "979-8-89176-251-0"
}

@inproceedings{
cai2026predictabilityreinforcementlearningdynamics,
title={On Predictability of Reinforcement Learning Dynamics for Large Language Models},
author={Cai Yuchen and Ding Cao and Xin Xu and Zijun Yao and Yuqing Huang and Benyi Zhang and Zhenyu Tan and Guiquan Liu and Junfeng Fang},
booktitle={The Fourteenth International Conference on Learning Representations},
year={2026},
url={https://openreview.net/forum?id=SdHmA6BYVJ}
}

@misc{cheng2026reasoning,
      title={Reasoning with Exploration: An Entropy Perspective}, 
      author={Daixuan Cheng and Shaohan Huang and Xuekai Zhu and Bo Dai and Wayne Xin Zhao and Zhenliang Zhang and Furu Wei},
      year={2025},
      eprint={2506.14758},
      archivePrefix={arXiv},
      primaryClass={cs.CL},
      url={https://arxiv.org/abs/2506.14758}, 
}

@inproceedings{
gandhi2025cognitivebehaviorsenableselfimproving,
title={Cognitive Behaviors that Enable Self-Improving Reasoners, or, Four Habits of Highly Effective {ST}aRs},
author={Kanishk Gandhi and Ayush K Chakravarthy and Anikait Singh and Nathan Lile and Noah Goodman},
booktitle={Second Conference on Language Modeling},
year={2025},
url={https://openreview.net/forum?id=QGJ9ttXLTy}
}

@inproceedings{agarwal2025unreasonableeffectivenessentropyminimization,
 author = {Agarwal, Shivam and Zhang, Zimin and Yuan, Lifan and Han, Jiawei and Peng, Hao},
 booktitle = {Advances in Neural Information Processing Systems},
 editor = {D. Belgrave and C. Zhang and H. Lin and R. Pascanu and P. Koniusz and M. Ghassemi and N. Chen},
 pages = {107150--107180},
 publisher = {Curran Associates, Inc.},
 title = {The Unreasonable Effectiveness of Entropy Minimization in LLM Reasoning},
 url = {https://proceedings.neurips.cc/paper_files/paper/2025/file/9a14cf2ee36cf2524f54d8f196a6734c-Paper-Conference.pdf},
 volume = {38},
 year = {2025}
}

@InProceedings{pmlr-v267-lin25j,
  title = 	 {Critical Tokens Matter: Token-Level Contrastive Estimation Enhances {LLM}’s Reasoning Capability},
  author =       {Lin, Zicheng and Liang, Tian and Xu, Jiahao and Liu, Qiuzhi and Wang, Xing and Luo, Ruilin and Shi, Chufan and Li, Siheng and Yang, Yujiu and Tu, Zhaopeng},
  booktitle = 	 {Proceedings of the 42nd International Conference on Machine Learning},
  pages = 	 {37906--37918},
  year = 	 {2025},
  editor = 	 {Singh, Aarti and Fazel, Maryam and Hsu, Daniel and Lacoste-Julien, Simon and Berkenkamp, Felix and Maharaj, Tegan and Wagstaff, Kiri and Zhu, Jerry},
  volume = 	 {267},
  series = 	 {Proceedings of Machine Learning Research},
  month = 	 {13--19 Jul},
  publisher =    {PMLR},
  url = 	 {https://proceedings.mlr.press/v267/lin25j.html}
}

@misc{han2025on,
      title={On the Role of Label Noise in the Feature Learning Process}, 
      author={Andi Han and Wei Huang and Zhanpeng Zhou and Gang Niu and Wuyang Chen and Junchi Yan and Akiko Takeda and Taiji Suzuki},
      year={2025},
      eprint={2505.18909},
      archivePrefix={arXiv},
      primaryClass={stat.ML},
      url={https://arxiv.org/abs/2505.18909}, 
}

@InProceedings{pmlr-v70-arpit17a,
  title = 	 {A Closer Look at Memorization in Deep Networks},
  author =       {Devansh Arpit and Stanis{\l}aw Jastrz{\k{e}}bski and Nicolas Ballas and David Krueger and Emmanuel Bengio and Maxinder S. Kanwal and Tegan Maharaj and Asja Fischer and Aaron Courville and Yoshua Bengio and Simon Lacoste-Julien},
  booktitle = 	 {Proceedings of the 34th International Conference on Machine Learning},
  pages = 	 {233--242},
  year = 	 {2017},
  editor = 	 {Precup, Doina and Teh, Yee Whye},
  volume = 	 {70},
  series = 	 {Proceedings of Machine Learning Research},
  month = 	 {06--11 Aug},
  publisher =    {PMLR},
  url = 	 {https://proceedings.mlr.press/v70/arpit17a.html}
}

@misc{mukherjee2026needadamsurprisinglystrong,
      title={Do We Need Adam? Surprisingly Strong and Sparse Reinforcement Learning with SGD in LLMs}, 
      author={Sagnik Mukherjee and Lifan Yuan and Pavan Jayasinha and Dilek Hakkani-Tür and Hao Peng},
      year={2026},
      eprint={2602.07729},
      archivePrefix={arXiv},
      primaryClass={cs.LG},
      url={https://arxiv.org/abs/2602.07729}, 
}

@inproceedings{jacot2018neural,
 author = {Jacot, Arthur and Gabriel, Franck and Hongler, Clement},
 booktitle = {Advances in Neural Information Processing Systems},
 editor = {S. Bengio and H. Wallach and H. Larochelle and K. Grauman and N. Cesa-Bianchi and R. Garnett},
 pages = {},
 publisher = {Curran Associates, Inc.},
 title = {Neural Tangent Kernel: Convergence and Generalization in Neural Networks},
 url = {https://proceedings.neurips.cc/paper_files/paper/2018/file/5a4be1fa34e62bb8a6ec6b91d2462f5a-Paper.pdf},
 volume = {31},
 year = {2018}
}

@inproceedings{chizat2019lazy,
 author = {Chizat, L\'{e}na\"{\i}c and Oyallon, Edouard and Bach, Francis},
 booktitle = {Advances in Neural Information Processing Systems},
 editor = {H. Wallach and H. Larochelle and A. Beygelzimer and F. d\textquotesingle Alch\'{e}-Buc and E. Fox and R. Garnett},
 pages = {},
 publisher = {Curran Associates, Inc.},
 title = {On Lazy Training in Differentiable Programming},
 url = {https://proceedings.neurips.cc/paper_files/paper/2019/file/ae614c557843b1df326cb29c57225459-Paper.pdf},
 volume = {32},
 year = {2019}
}

@InProceedings{malladi2023kernel,
  title = 	 {A Kernel-Based View of Language Model Fine-Tuning},
  author =       {Malladi, Sadhika and Wettig, Alexander and Yu, Dingli and Chen, Danqi and Arora, Sanjeev},
  booktitle = 	 {Proceedings of the 40th International Conference on Machine Learning},
  pages = 	 {23610--23641},
  year = 	 {2023},
  editor = 	 {Krause, Andreas and Brunskill, Emma and Cho, Kyunghyun and Engelhardt, Barbara and Sabato, Sivan and Scarlett, Jonathan},
  volume = 	 {202},
  series = 	 {Proceedings of Machine Learning Research},
  month = 	 {23--29 Jul},
  publisher =    {PMLR},
  url = 	 {https://proceedings.mlr.press/v202/malladi23a.html}
}

@misc{parthasarathi2025grpolambdacreditassignmentimproves,
      title={GRPO-$\lambda$: Credit Assignment improves LLM Reasoning}, 
      author={Prasanna Parthasarathi and Mathieu Reymond and Boxing Chen and Yufei Cui and Sarath Chandar},
      year={2025},
      eprint={2510.00194},
      archivePrefix={arXiv},
      primaryClass={cs.LG},
      url={https://arxiv.org/abs/2510.00194}, 
}

@misc{he2025justrlscaling15bllm,
      title={JustRL: Scaling a 1.5B LLM with a Simple RL Recipe}, 
      author={Bingxiang He and Zekai Qu and Zeyuan Liu and Yinghao Chen and Yuxin Zuo and Cheng Qian and Kaiyan Zhang and Weize Chen and Chaojun Xiao and Ganqu Cui and Ning Ding and Zhiyuan Liu},
      year={2025},
      eprint={2512.16649},
      archivePrefix={arXiv},
      primaryClass={cs.CL},
      url={https://arxiv.org/abs/2512.16649}, 
}

@misc{li2026outcomegroundedadvantagereshapingfinegrained,
      title={Outcome-Grounded Advantage Reshaping for Fine-Grained Credit Assignment in Mathematical Reasoning}, 
      author={Ziheng Li and Liu Kang and Feng Xiao and Luxi Xing and Qingyi Si and Zhuoran Li and Weikang Gong and Deqing Yang and Yanghua Xiao and Hongcheng Guo},
      year={2026},
      eprint={2601.07408},
      archivePrefix={arXiv},
      primaryClass={cs.CL},
      url={https://arxiv.org/abs/2601.07408}, 
}
